%% file: neurips_2026.tex
\documentclass{article}

\PassOptionsToPackage{most}{tcolorbox}
\usepackage[final]{acl}

\usepackage{times}
\usepackage{latexsym} 
\usepackage[utf8]{inputenc}
\usepackage[T1]{fontenc}
\usepackage{amsmath,amssymb}
\usepackage{amsfonts}
\usepackage{booktabs}
\usepackage{microtype}
\usepackage{inconsolata}
\usepackage{graphicx}
\usepackage{wrapfig}
\usepackage{enumitem}
\usepackage{pifont}
\usepackage{nicefrac}
\usepackage{listings}

\usepackage[utf8]{inputenc} 
\usepackage[T1]{fontenc}    
\usepackage{hyperref}       
\usepackage{url}            
\usepackage{booktabs}       
\usepackage{amsfonts}       
\usepackage{nicefrac}       
\usepackage{microtype}      
\usepackage{xcolor}         
\usepackage{placeins}
\usepackage[most]{tcolorbox}
\tcbuselibrary{listings,breakable}
\usepackage{multirow}
\usepackage{longtable}
\usepackage{array} 
\usepackage{graphicx}
\usepackage{wrapfig}
\usepackage{makecell}
\usepackage{pifont}
\usepackage{array}
\usepackage{caption}
\usepackage{subcaption}

\usepackage{times}
\usepackage{latexsym} 
\usepackage[utf8]{inputenc}
\usepackage[T1]{fontenc}
\usepackage{amsmath,amssymb}
\usepackage{amsfonts}
\usepackage{booktabs}
\usepackage{microtype}
\usepackage{inconsolata}
\usepackage{graphicx}
\usepackage{wrapfig}
\usepackage{enumitem}
\usepackage{pifont}
\usepackage{nicefrac}
\usepackage{listings}

\usepackage{enumitem}
\usepackage{multirow}
\usepackage{wrapfig}
\usepackage{tabularx}
\usepackage{svg}
\usepackage[utf8]{inputenc} 
\usepackage[T1]{fontenc}    
\usepackage{hyperref}       
\usepackage{url}            
\usepackage{booktabs}       
\usepackage{amsfonts}       
\usepackage{nicefrac}       
\usepackage{microtype}      
\usepackage{xcolor}         
\usepackage{placeins}
\usepackage{caption}
\usepackage[most]{tcolorbox}
\usepackage{longtable}
\usepackage{array}
\usepackage{placeins}
\tcbuselibrary{listings,breakable}
\tcbuselibrary{skins,breakable}

\tcbuselibrary{listings, breakable, skins}
\definecolor{DeepPurple}{HTML}{673AB7}
\definecolor{LighterGray}{HTML}{FAFAFA}
\definecolor{CaseOrange}{HTML}{F57C00}

\newcommand{\answerTODO}[1][]{\textcolor{red}{\bf [TODO]}}
\newcommand{\justificationTODO}[1][]{\textcolor{red}{\bf [TODO]}}

\DeclareRobustCommand{\cmark}{\textcolor{green!60!black}{\ding{51}}}
\DeclareRobustCommand{\xmark}{\textcolor{red!70!black}{\ding{55}}}

\definecolor{formatblue}{RGB}{51,51,144}
\definecolor{formatlavender}{RGB}{224,224,255}
\definecolor{formatgray}{RGB}{242,242,242}
\definecolor{formatblack}{RGB}{25,25,25}
\definecolor{formatolive}{RGB}{142,140,90}
\definecolor{formatcream}{RGB}{250,248,236}
\definecolor{formatteal}{RGB}{0,94,96}

\newtcblisting{schemaformatbox}[1]{
  enhanced,
  breakable,
  boxrule=0.2pt,
  arc=2pt,
  colback=formatcream,
  colframe=formatolive,
  coltitle=white,
  colbacktitle=formatolive,
  coltext=formatteal,
  title=#1,
  fonttitle=\bfseries,
  listing only,
  listing options={
    basicstyle=\ttfamily\normalsize\color{formatteal},
    breaklines=true,
    columns=fullflexible
  }
}

\newtcblisting{schemabox}[1]{
  enhanced,
  breakable,
  boxrule=0.2pt,
  colback=gray!5,
  colframe=black,
  coltitle=white,
  colbacktitle=black,
  title=#1,
  fonttitle=\bfseries,
  listing only,
  listing options={
    basicstyle=\ttfamily\normalsize,
    breaklines=true,
    columns=fullflexible
  }
}

\newtcblisting{promptformatbox}[1]{
  enhanced,
  breakable,
  boxrule=0.2pt,
  arc=2pt,
  colback=formatgray,
  colframe=formatblue,
  coltitle=black,
  colbacktitle=formatlavender,
  coltext=formatblack,
  title=#1,
  fonttitle=\bfseries,
  listing only,
  listing options={
    basicstyle=\ttfamily\normalsize\color{formatblack},
    breaklines=true,
    columns=fullflexible
  }
}

\title{SMH-Bench: Benchmarking LLM Agents for Environment-Grounded Reasoning and Action in Smart Homes
}

\author{%
  \textbf{Kuan Li$^{1,*}$}\quad
  \textbf{Shuo Zhang$^{1,2,*,\diamond}$}\quad
  \textbf{Huacan Wang$^{1,*,\dagger,\ddagger}$}\quad
  \textbf{Fangzhou Yu$^{1,3,\diamond}$}\\
  \textbf{Zecheng Sheng$^{1,4,\diamond}$}\quad
  \textbf{Yi Gu$^{1}$}\quad
  \textbf{Weipeng Ming$^{1}$}\quad
  \textbf{Lei Xue$^{1}$}\quad
  \textbf{Chen Liu$^{1}$}\\
  \textbf{Sen Hu$^{5}$}\quad
  \textbf{Ronghao Chen$^{5}$}\quad
  \textbf{Siyue Lin$^{1}$}\quad
  \textbf{Yuqing Hou$^{1}$}\quad
  \textbf{Xiaofeng Mou$^{1}$}\quad
  \textbf{Yi Xu$^{1,\dagger}$}\\[6pt]
  $^{1}$Midea Group\quad
  $^{2}$Beijing University of Posts and Telecommunications\\
  $^{3}$Donghua University\quad
  $^{4}$The University of Sydney\quad
  $^{5}$Peking University\\[3mm]
  $^{*}$Equal Contribution,\quad
  $^{\dagger}$Corresponding Author,\quad
  $^{\ddagger}$Project Leader.\\[2mm]
  $^{\diamond}$Work completed during an internship at Midea AI Research Center.
}

\begin{document}

\maketitle

\begin{abstract}
  Smart homes are evolving toward complex state-dependent living environments, requiring Large Language Models (LLMs) to reason over user intent, preferences, and multi-device interactions. However, existing smart-home benchmarks often focus on static instruction-to-API mapping or limited simulations, failing to evaluate whether LLMs can reason, interact, and act reliably in realistic household scenarios. To address these limitations, we introduce SMH-Bench, a comprehensive benchmark for evaluating LLMs in smart-home environments. Built upon HomeEnv, an executable and verifiable smart-home simulator, SMH-Bench contains 1,100 high-quality tasks spanning 7 categories and 22 fine-grained subcategories. It further stratifies tasks across simple, medium and complex homes, ranging from small apartments to dense multi-room environments with 135 devices. Experiments show that although frontier LLMs achieve strong performance on explicit control and query tasks, they still exhibit significant weaknesses in automation task scheduling, ambiguity handling and personalized reasoning, especially as home complexity increases. We hope SMH-Bench will facilitate the development of more reliable, context-aware, and practically deployable smart-home agents.

  \vspace{0.8em}
  \noindent\textbf{Keywords:} Smart Home Benchmark, LLM Agent, Home Simulation, Interactive Evaluation

  \vspace{0.4em}
  \noindent\textbf{Correspondence:} 
  \href{mailto:wanghuacan17@mails.ucas.ac.cn}{Huacan Wang},
  \href{mailto:xuyi42@midea.com}{Yi Xu} \\

  \vspace{0.2em}
  \noindent\textbf{Data and Code:} Coming soon.
  
\end{abstract}

\vspace{5mm}

\input{sections/1.introduction}
\input{sections/2.related_work}
\input{sections/3.benchmark}
\input{sections/4.experiments}
\input{sections/5.conclusion}



\bibliography{references}

\newpage
\appendix
\input{appendix/mha_state_space1}
\input{appendix/task_instruction_generation_rules_backup4}
\input{appendix/task_taxonomy}

\input{appendix/implementation_details_for_the_two_evaluation_settings_backup7}
\input{appendix/error_distribution_breakdown}
\input{appendix/detailed_model_results}
\input{appendix/prompt_templates}

%

\end{document}

%% file: sections/1.introduction.tex
\section{Introduction}



Smart homes are evolving toward increasingly complex and personalized user demands, requiring systems to understand natural language instructions, infer implicit user intent, and coordinate actions across multiple devices and rooms. Recently, large language model (LLM)-driven agents have demonstrated significant potential in intent understanding~\cite{wang2025learning, zhang2025future}, tool invocation~\cite{liu2025toolace, patil2025bfcl}, and multi-step decision-making~\cite{shridhar2021alfworld, yao2022webshop, kim2025reflact}, making them a promising foundation for complex smart home tasks~\cite{King2024Sasha, Rivkin2025AIoT, yin2025harmony}. Consequently, rigorously evaluating these agents in authentic, complex smart home settings has emerged as a critical challenge.

While current evaluation benchmarks have fueled progress in this field, they still exhibit notable limitations. Many benchmarks~\cite{King2024Sasha,li2025homebench} formulate tasks as static instruction-to-API mappings, which evaluates basic comprehension but ignores dynamic home states and fails to verify success through environmental feedback. Recent works~\cite{Rivkin2025AIoT,seo2026simuhome} further introduce state-aware simulations to ground evaluations in interactive environments. However, their tasks remain confined to limited scenarios like simple queries and device control, leaving the complex, continuous, and personalized demands of real-world interactions underexplored. In practice, user instructions are rarely isolated, fully specified single-turn commands. Instead, they frequently span multiple devices and rooms, requiring agents to disambiguate intent based on real-time states, dialogue context, and user preferences. Moreover, as home ecosystems scale, agents must perform complex querying, filtering, and compositional control across extensive device networks. Ultimately, existing benchmarks still fall short of comprehensively evaluating whether agents can understand, reason, remember, and act in realistic homes.

\begin{figure}[ht]
  \centering
  \includegraphics[width=0.9\textwidth]{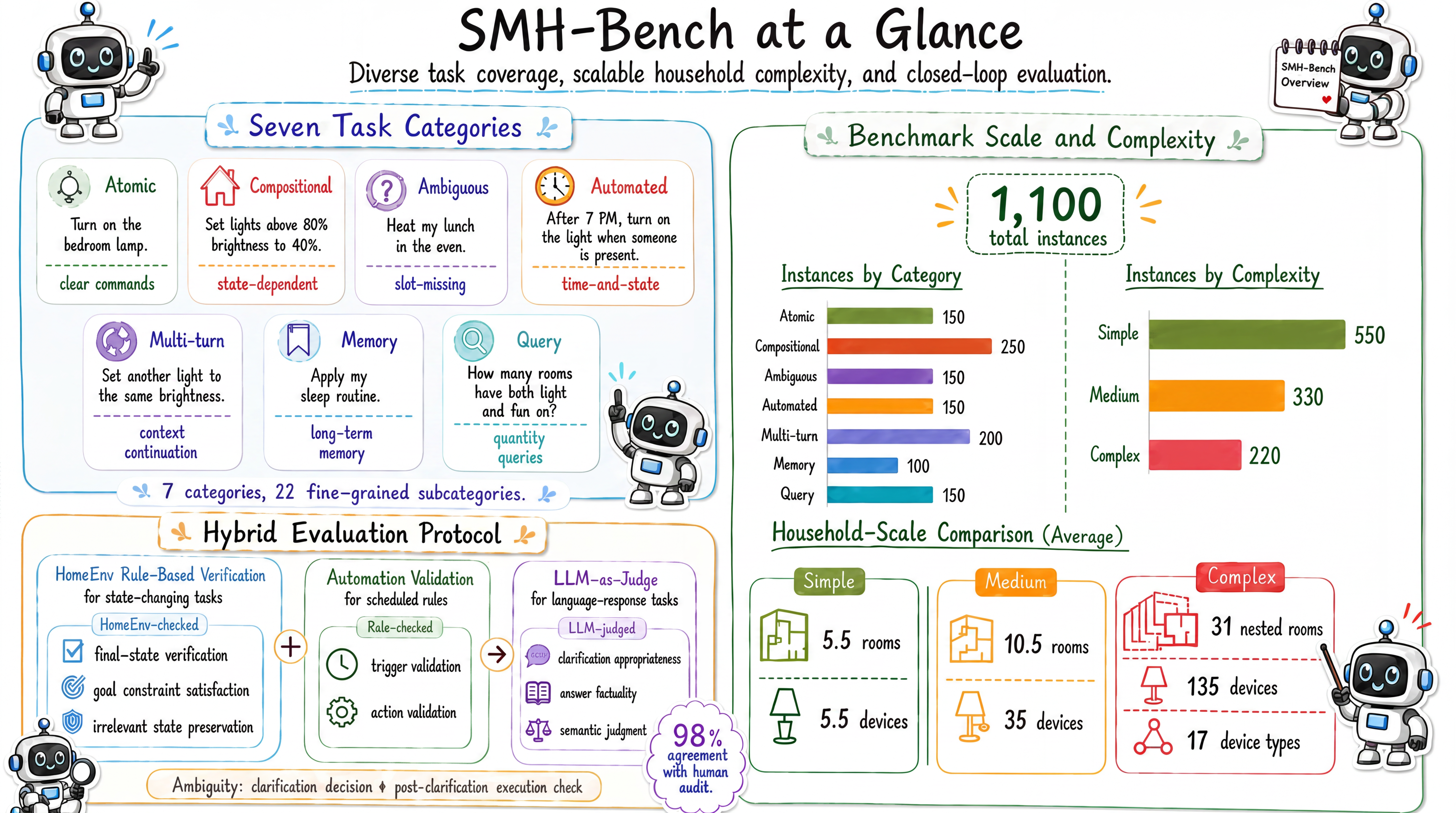}
  \caption{Overview of SMH-Bench. SMH-Bench evaluates agents with 1,100 human-audited tasks across seven capability categories, scalable household complexity, and a hybrid evaluation protocol.
}
\label{fig:first}
\end{figure}

To address these limitations, we introduce SMH-Bench, a more comprehensive benchmark for evaluating smart-home agents in realistic household settings. With the goal of defining, executing, and verifying diverse task types within a unified framework, we design HomeEnv, a stateful, executable, and verifiable smart home environment, as its foundational infrastructure. HomeEnv models the home as a continuously evolving state, encompassing rooms, devices, and their dynamic attributes. Unlike static API-based evaluations, HomeEnv allows agents to interact with executable home states and verifies task success based on the actual state transitions caused by agent actions. This design enables flexible evaluation of heterogeneous tasks without relying on rigid reference action sequences. Table~\ref{tab:task_comparison} compares SMH-Bench with existing smart-home benchmarks.

Building upon HomeEnv, we introduce a task hierarchy designed to evaluate the core reasoning and execution capabilities of smart-home agents. SMH-Bench covers diverse dimensions including linguistic robustness, state-driven reasoning, multi-turn interaction, and personalized memory. For example, a user might remark that ``it feels a bit chilly, so please adjust the heating in the rooms I usually use to the same temperature as last night.'' Fulfilling this requires the agent to query real-time environmental variables, retrieve preferences from historical dialogue, and identify the user's frequently occupied rooms. In total, we construct 1,100 tasks across 7 categories, spanning household scales from simple apartments with only 8 devices to complex environments with 135 devices, as detailed in Figure~\ref{fig:first}.

We comprehensively evaluate 13 representative LLMs using SMH-Bench, revealing a persistent capability gap in authentic smart home interactions. Although top models like Gemini-3.1-Pro achieve peak success rates over 85\%, performance remains highly inconsistent across task complexities. While agents reliably handle explicit atomic controls, they experience severe performance degradation in complex scenarios requiring ambiguous intent resolution, compositional multi-device coordination, state-driven automation, and personalized memory utilization. 

%% file: sections/2.related_work.tex
\section{Related Work}

\textbf{LLM Agents in Smart Homes.} LLM-based smart-home agents have broadened smart-home interaction from command parsing to context-grounded decision making. Earlier studies on smart speakers and home assistants examined long-term use, proactive interaction, and privacy risks~\cite{mckie2022longterm,zargham2022proactive,lutz2021privacy}, while recent LLM-based systems study several concrete capabilities. Sasha~\cite{King2024Sasha} and SAGE~\cite{sage2023} focus on ambiguous goal reasoning and interaction with household context and device APIs. ChatIoT~\cite{chatiot2024} and AwareAuto~\cite{awareauto2024} translate natural-language instructions into automation rules, and IoTGPT~\cite{iotgpt2026}, HomeLLaMA~\cite{homellama2025}, and Harmony~\cite{yin2025harmony} further consider memory, on-device execution, latency, and privacy. These studies show that LLM-based smart-home agents are increasingly defined not by a single function such as command parsing or device control, but by the coordination of multiple capabilities, including intent grounding, device/API interaction, automation, memory, personalization, and reasoning over household-specific constraints.


\textbf{Evaluation and Benchmarks in Smart Homes.} Smart-home benchmarks have evolved from activity recognition and embodied household tasks toward LLM-based assistant evaluation. CASAS~\cite{cook2012casas} and ARAS~\cite{alemdar2013aras} focus on activity recognition, while VirtualHome~\cite{puig2018virtualhome}, ALFRED~\cite{shridhar2020alfred}, TEACh~\cite{padmakumar2022teach}, BEHAVIOR~\cite{srivastava2021behavior}, and ReALFRED~\cite{kim2024realfred} emphasize navigation and physical manipulation in household environments. Recent smart-home benchmarks are closer to our setting: HomeBench~\cite{li2025homebench} evaluates valid and invalid device-control instructions, SmartHome-Bench~\cite{smarthomebench2025} and SmartBench~\cite{smartbench2026} study anomaly and safety-related scenarios, and SimuHome~\cite{seo2026simuhome} introduces executable temporal simulation. Compared with these efforts, SMH-Bench expands the evaluation scope from isolated device control or safety scenarios to a broader set of environment-grounded tasks involving compositional control, ambiguity resolution, automation, dialogue context, memory, and state queries, while using HomeEnv to execute agent actions and verify final-state correctness, preservation of irrelevant states, and task-specific outcomes such as automation conditions, clarification decisions, and query answers across homes of different complexity.

\begin{table}[t]
\centering
\small
\setlength{\tabcolsep}{4pt}
\caption{Comparing SMH-Bench to existing benchmarks for smart-home evaluation.}
\begin{tabular}{lccccccc}
\toprule
\textbf{Benchmark}
& \thead{Device\\Query}
& \thead{Automation}
& \thead{Clarify\\Necessity}
& \thead{Spoken /\\Correction}
& \thead{Multi-turn\\Dialogue}
& \thead{Personalized\\Memory}
& \thead{Dataset\\Size} \\
\midrule
HomeBench & \xmark & \xmark & \xmark & \xmark & \xmark & \xmark & 170K \\
SimuHome & \cmark & \cmark & \cmark & \xmark & \xmark & \xmark & 600 \\
\textbf{SMH-Bench} & \cmark & \cmark & \cmark & \cmark & \cmark & \cmark & 1,100 \\
\bottomrule
\end{tabular}
\label{tab:task_comparison}
\end{table}

%% file: sections/3.benchmark.tex
\section{SMH-Bench}

SMH-Bench aims to provide a comprehensive and realistic evaluation benchmark for LLM-driven agents in smart-home environments. 
To ground agent evaluation in concrete household contexts, we build a simulated smart-home environment, named HomeEnv, which models room layouts, device configurations, and dynamic device states. 
In this section, we first formalize the interaction task for smart-home agents and then describe the design of HomeEnv. 
We further present the benchmark taxonomy, data construction pipeline, dataset statistics, and evaluation protocol. 
Figure~\ref{fig:dataset_pipeline} illustrates the construction process of SMH-Bench.

\subsection{Task Formulation}

We formulate smart-home agent evaluation as an environment-grounded interaction task. 
At interaction turn $t$, the smart home is represented as $\mathcal{H}_t=(\mathcal{R},\mathcal{D},\phi,\mathcal{X}_t,\mathcal{S})$, where $\mathcal{R}$ is the set of rooms, $\mathcal{D}$ is the set of devices, $\phi(d_j)\in\mathcal{R}$ maps each device $d_j$ to its located room, $\mathcal{X}_t(d_j)$ denotes the dynamic attributes of $d_j$, and $\mathcal{S}(d_j)$ denotes its callable services with corresponding arguments. 
Given a user instruction $u$, the agent needs to ground the instruction in the current home state and decide the appropriate response.

For executable control tasks, the agent produces one or more service calls. 
A single action is denoted as $a=[d_j.s_k(\mathbf{p})]$, where $d_j$ is the target device, $s_k\in\mathcal{S}(d_j)$ is a callable service, and $\mathbf{p}$ denotes the service arguments. 
Executing an action updates the environment through $\mathcal{H}_{t+1}=T(\mathcal{H}_t,a)$. 
For tasks involving multiple operations, the agent outputs an action sequence $A=(a_1,\ldots,a_n)$, which induces a sequence of state transitions from the initial state $\mathcal{H}_0$ to the final state $\mathcal{H}_n$.

To cover realistic household interactions, each benchmark instance is defined as $\tau=(\mathcal{H}_0,u,\mathcal{C},\mathcal{M},g)$, where $\mathcal{C}$ is the dialogue history, $\mathcal{M}$ is user memory, and $g$ is the task-specific goal. 
Depending on the task type, $g$ may specify target device states, expected query answers, clarification requirements, or scheduled execution conditions. 
Accordingly, the agent output can be executable service calls, a natural-language answer, a clarification question, or a scheduled rule. 
A task is considered successful if the agent output satisfies $g$ under the current environment: executable tasks are verified by whether the final state $\mathcal{H}_n$ satisfies the target constraints while preserving irrelevant device states, whereas non-executable responses are evaluated by task-specific criteria described in Section~\ref{subsec:eval}.

\subsection{HomeEnv: Simulated Smart-Home Environment}

HomeEnv is a lightweight simulated smart-home environment designed to support executable agent interactions in realistic household settings. 
It serves as the environment foundation of SMH-Bench by providing structured home states, configurable device abstractions, and standardized query and control interface 

\textbf{Smart-Home State Space.}
HomeEnv represents a home as a structured environment composed of rooms and devices. 
Rooms define the spatial organization of the household and may contain nested structures, such as a bathroom inside a bedroom. 
Devices are assigned to rooms and described by their type and location. 
Each device is further represented through attributes and services: attributes describe current device states, such as power status, brightness, target temperature, humidity, or operation mode, while services define executable operations, such as turning a device on or off, setting a temperature, adjusting brightness, or switching modes. 
The full schema of rooms, devices, attributes, and services is provided in Appendix~\ref{app:HomeEnv_state_space}.

\textbf{Device Operation Engine.}
The operation engine updates the home state by executing structured actions issued by the agent. 
For each device-control action, the agent specifies the target device, the intended operation, and its parameters. 
HomeEnv checks whether the requested operation is valid for the target device, including whether the operation exists and whether the provided parameters satisfy predefined type, range, and option constraints. 
If the action is valid, the engine applies the operation and updates the corresponding device attributes; otherwise, the invalid operation is rejected and the home state remains unchanged.

\textbf{Agent-Environment Interface.}
Agents interact with HomeEnv through structured query and control actions. 
The query interface allows agents to inspect the simulated home before acting, such as retrieving the room layout, finding devices in a specific room, listing devices of a certain type, checking their current states, or obtaining the operations that can be applied to them. 
The control interface allows agents to execute device operations through structured action objects. 
After each interaction, HomeEnv returns an observation containing the requested query result or the execution status of the control action. 
This standardized interface enables agents to perform perception and action within the same simulated home state.

\subsection{Benchmark Construction}

\begin{figure}[t]
  \centering
  \includegraphics[width=0.9\textwidth]{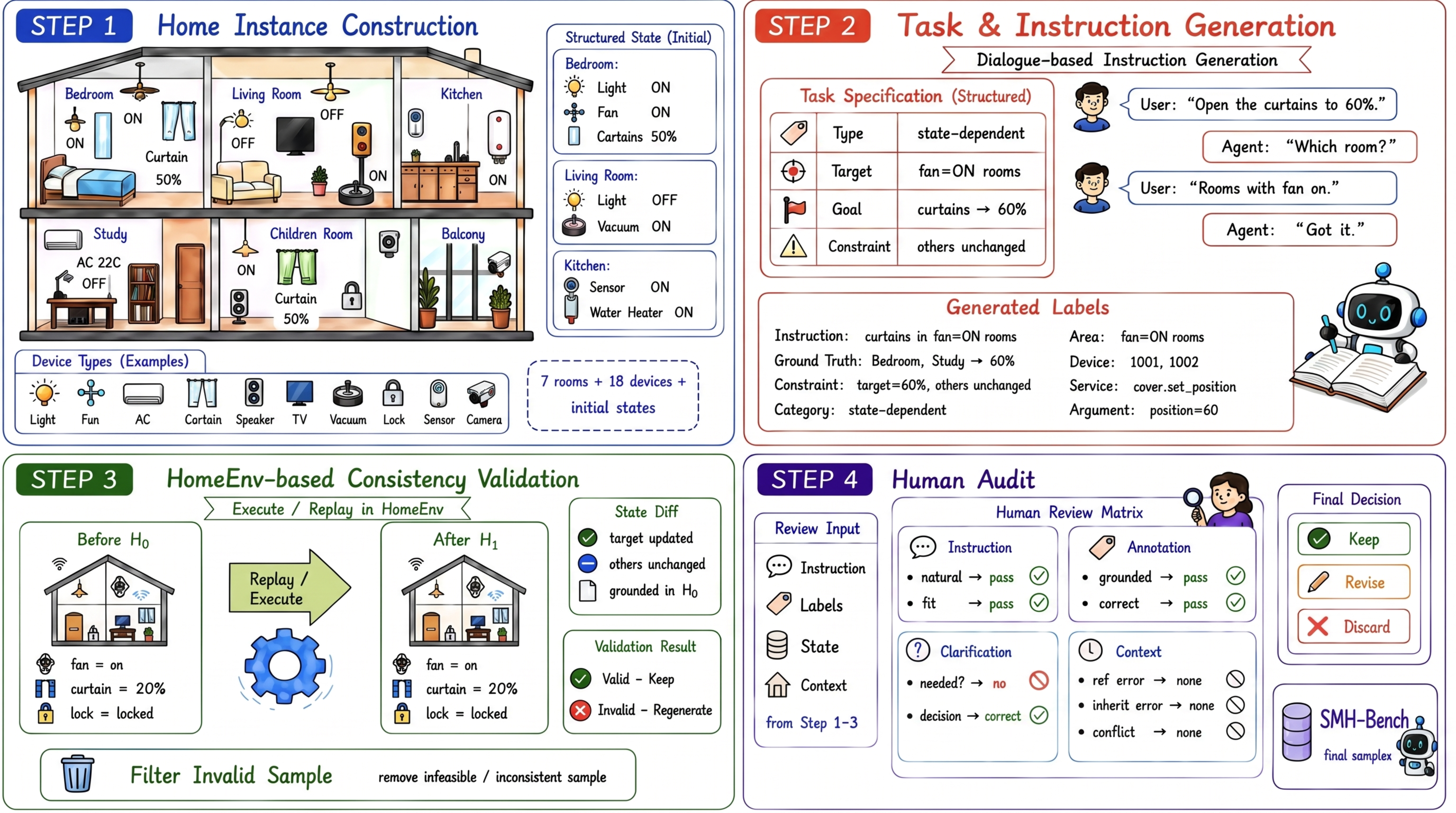}
  \caption{SMH-Bench construction pipeline. Each instance is built through four stages: home instance construction, task and instruction generation from structured home states, HomeEnv-based consistency validation, and final human audit before inclusion in the benchmark.}
  \label{fig:dataset_pipeline}
\end{figure}

\subsubsection{Task Taxonomy}
\label{sec:TC}

To evaluate smart-home agents beyond device-type-specific command execution, SMH-Bench organizes tasks by environment-grounded capabilities. The taxonomy contains seven primary categories and 22 fine-grained subcategories. Atomic Control (TC1) covers single-intent operations under clear, colloquial, or noisy wording, while Compositional Control (TC2) evaluates multi-device, state-dependent, room-dependent, and selection-based execution. For example, ``\textit{except for the hallway, turn on all fans and air conditioners in rooms whose humidity is above 60\%}'' requires the agent to first identify rooms satisfying a sensor predicate and then apply actions only to devices located in those rooms.

Beyond explicit execution, SMH-Bench evaluates whether agents can handle ambiguity, automation, dialogue context, memory, and state queries. Ambiguous Intent (TC3) covers subjective complaints, scene-level goals, and slot-missing requests; for instance, ``\textit{the dining room feels like a steamer}'' requires mapping a discomfort complaint to a feasible cooling intervention or asking for clarification when the state is underspecified. Automated Task Scheduling (TC4) evaluates persistent time- or state-triggered rules, such as turning on hallway and staircase lights at a specified brightness when motion is detected in selected rooms. Context-Aware Multi-turn Interaction (TC5) tests whether the current turn inherits, revises, or replaces previous operations, while Personalized Memory (TC6) requires applying short- or long-term user preferences, such as setting another air conditioner ``\textit{the same way as before}''. Environment-Grounded Query (TC7) evaluates non-mutating inspection and aggregation over the current home state. Together, these categories stress-test state grounding, target selection, clarification, rule construction, dialogue tracking, memory use, query reasoning, and irrelevant-state preservation. Full subcategory definitions, representative examples, and evaluation targets are provided in Appendix~\ref{sec:appendix_examples}.

\subsubsection{Data Construction Pipeline}
SMH-Bench is constructed with an environment-first pipeline, as shown in Figure~\ref{fig:dataset_pipeline}. We first instantiate concrete home states and then derive tasks, instructions, contexts, and labels from these states. This design makes each instance directly verifiable against HomeEnv.

\textbf{Step1: Home Instance Construction.} 
We construct homes at three complexity levels. 
Simple and medium instances use independently generated homes to increase layout and device diversity, whereas complex instances share a large dense home with nested rooms and rich device deployments to support multi-room reasoning and state comparison. 

\textbf{Step2: Task and Instruction Generation.} 
For each home, GPT-5 first produces a structured task specification conditioned on the home state, task category, and generation constraints. 
The specification records the target devices, desired states, comparison rules, contextual dependencies, or expected answers needed for evaluation. 
GPT-5 then verbalizes the specification into a natural user instruction, optionally incorporating dialogue history or user memory for multi-turn and personalized tasks. 
Detailed generation rules and examples are provided in Appendix~\ref{app:task_instruction_generation}.

\textbf{Step3: HomeEnv-based Consistency Validation.} 
We automatically validate generated instances with HomeEnv before human review. 
For executable tasks, HomeEnv checks device existence, service availability, argument validity, and whether the reference operations reach the intended final state. 
For non-executable tasks, it recomputes query answers, verifies scheduling triggers and actions, and checks whether ambiguous requests indeed require clarification under the current state. 
Instances with state-label inconsistencies are revised or removed.

\textbf{Step4: Human Audit.} 
Finally, human annotators review the remaining instances for instruction naturalness, category alignment, and context or memory conflicts. 
This final audit targets semantic and pragmatic errors that are difficult to catch through simulator checks alone.

\begin{wrapfigure}{r}{0.55\textwidth}
  \centering
  \includegraphics[width=0.55\textwidth]{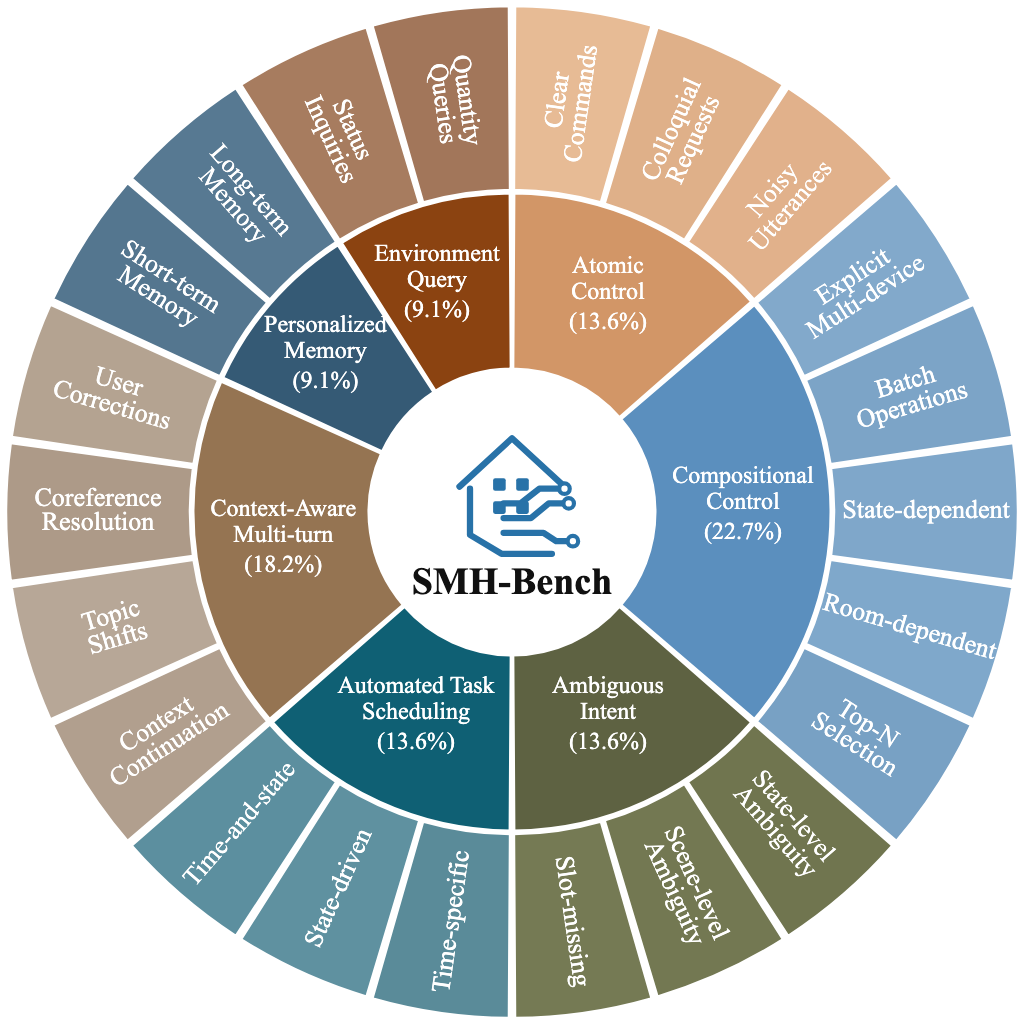}
  \caption{Task distribution of SMH-Bench. The benchmark organizes 1,100 instances into seven primary categories and 22 fine-grained subcategories.}
  \label{fig:task_pie_chart}
\end{wrapfigure}

\subsubsection{Data Statistics and Analysis}
\label{subsubsec:stat}

SMH-Bench contains 1,100 human-audited evaluation instances across the taxonomy defined in Section~\ref{sec:TC}. As shown in Figure~\ref{fig:task_pie_chart}, the dataset is not uniformly distributed across categories. Compositional Control and Context-Aware Multi-turn Interaction form the two largest groups, with 250 and 200 instances, respectively. Atomic Control, Ambiguous Intent, Automated Task Scheduling, and Environment-Grounded Query each contain 150 instances, while Personalized Memory contains 100 instances. This allocation reflects the benchmark design choice to retain broad category coverage while assigning more evaluation mass to interaction settings that involve richer environment dependence.

SMH-Bench further stratifies instances by environmental complexity into simple, medium, and complex settings, with 550, 330, and 220 instances, respectively. As illustrated in Figure~\ref{fig:first}, these settings differ in spatial scale and device density. Simple environments model small apartments with an average of 5.5 rooms and 5.5 devices, while medium environments model multi-room homes with an average of 10.5 rooms and 35 devices. The complex setting uses a dense hierarchical home with 31 nested rooms and 135 devices across 17 device types. This stratification provides a controlled axis for analyzing how agent performance changes as the observable home state becomes larger and more distractor-rich.

\subsection{Evaluation Protocol}
\label{subsec:eval}

SMH-Bench adopts a hybrid evaluation protocol that combines rule-based verification with LLM-as-a-Judge assessment. We use rule-based evaluation for tasks that can be verified in HomeEnv, and LLM-as-a-Judge evaluation for tasks that require semantic assessment.

\textbf{Rule-Based Verification.} For tasks involving physical environment changes, the HomeEnv simulator objectively verifies outcomes by comparing the post-execution state against the ground-truth goal. This process enforces an \textit{irrelevant state preservation constraint} to ensure that devices not explicitly targeted remain strictly unchanged. Notably, for automated task scheduling (TC4), we implement a dual-rule validation mechanism: (1) \textit{Trigger Validation}, where temporal triggers are verified by matching Cron trigger times, and state-driven triggers are evaluated by ensuring they trigger under the exact same device states as the ground truth; and (2) \textit{Action Validation}, which verifies the resulting operations using the aforementioned state comparison.

\textbf{LLM-as-a-Judge Assessment.} For tasks where the agent's output is a natural language response, such as ambiguous intent (TC3) and environment-grounded queries (TC7), we utilize GPT-5 to assess the appropriateness of clarification requests and the factual accuracy of the answers. To guarantee the reliability of this process, we conducted a blind human audit on 200 randomly sampled instances, achieving a 98\% agreement rate between the LLM judge and human experts.

%% file: sections/4.experiments.tex
\section{Experiments}

\begin{table}[t]
\centering
\small
\caption{Performance comparison under EIA and DR settings. TC1--TC7 are defined in Section~\ref{sec:TC}. For each model, success rates on individual capabilities and the overall average are reported in percentages. Token usage is reported in thousands. $\Delta$ Avg. measures the average success-rate difference between EIA and DR, where positive values indicate that EIA improves over DR.}
\label{tab:main_results}
\resizebox{\textwidth}{!}{
\begin{tabular}{@{}llcccccccccc@{}}
\toprule
\multirow{2}{*}{Model}
& \multirow{2}{*}{Setting}
& \multicolumn{8}{c}{Capability (Success Rate, \%)}
& \multirow{2}{*}{\shortstack{Tokens (k)\\($\downarrow$)}}
& \multirow{2}{*}{\shortstack{$\Delta$ Avg.\\(EIA--DR)}} \\
\cmidrule(lr){3-10}
& & TC1 & TC2 & TC3 & TC4 & TC5 & TC6 & TC7 & Avg. & & \\
\midrule
\multicolumn{12}{c}{\textit{Open-source Models}} \\
\midrule
\multirow{2}{*}{Qwen3.5-4B}
& DR & 68.7 & 35.6 & 44.7 & 24.0 & 48.0 & 39.0 & 75.0 & 45.9 & 1.2 & \multirow{2}{*}{$+10.6$} \\
& EIA & 69.3 & 45.6 & 72.0 & 21.3 & 65.0 & 57.0 & 76.0 & 56.5 & 8.3 & \\
\multirow{2}{*}{Qwen3.5-9B}
& DR & 73.3 & 38.0 & 47.3 & 25.3 & 51.5 & 47.0 & 77.0 & 49.2 & 1.4 & \multirow{2}{*}{$+15.1$} \\
& EIA    & 76.7 & 56.0 & 69.3 & 36.0 & 74.0 & 65.0 & 81.0 & 64.3 & 7.4 & \\
\multirow{2}{*}{Qwen3.5-397B}
& DR & 92.7 & 72.0 & 76.7 & 60.0 & 85.0 & 75.0 & 84.0 & 77.6 & 6.0 & \multirow{2}{*}{$+0.2$} \\
& EIA    & 88.0 & 68.8 & 86.7 & 69.3 & 74.5 & 77.0 & 92.0 & 77.8 & 7.3 & \\
\multirow{2}{*}{DeepSeek-V3.2}
& DR & 79.3 & 60.8 & 65.3 & 52.0 & 80.0 & 68.0 & 89.0 & 69.5 & 1.4 & \multirow{2}{*}{$+6.0$} \\
& EIA    & 87.3 & 52.8 & 85.3 & 64.0 & 85.0 & 74.0 & 94.0 & 75.5 & 12.9 & \\
\multirow{2}{*}{MiniMax-M2.7}
& DR & 81.3 & 55.6 & 60.7 & 38.7 & 32.5 & 32.0 & 63.0 & 51.8 & 2.5 & \multirow{2}{*}{$+0.3$} \\
& EIA    & 57.3 & 23.6 & 52.0 & 42.0 & 41.0 & 27.0 & 90.0 & 52.1 & 4.7 & \\
\multirow{2}{*}{GLM-5}
& DR & 80.7 & 50.0 & 49.3 & 52.0 & 71.0 & 39.0 & 78.0 & 59.7 & 3.1 & \multirow{2}{*}{$+10.9$} \\
& EIA    & 88.7 & 65.6 & 83.3 & 42.7 & 76.0 & 51.0 & 87.0 & 70.6 & 3.6 & \\
\midrule
\multicolumn{12}{c}{\textit{Closed-source Models}} \\
\midrule
\multirow{2}{*}{Qwen3.5-Plus}
& DR & 79.3 & 51.6 & 51.3 & 50.0 & 74.0 & 71.0 & 69.0 & 62.6 & 5.2 & \multirow{2}{*}{$+7.4$} \\
& EIA    & 91.3 & 57.2 & 76.7 & 42.7 & 79.5 & 65.0 & 87.0 & 70.0 & 8.3 & \\
\multirow{2}{*}{GPT-5.4-Mini}
& DR & 80.7 & 66.0 & 61.3 & 45.3 & 79.0 & 62.0 & 89.0 & 68.6 & 4.9 & \multirow{2}{*}{$+6.7$} \\
& EIA    & 89.3 & 68.8 & 86.0 & 50.7 & 81.0 & 69.0 & 86.0 & 75.3 & 3.3 & \\
\multirow{2}{*}{GPT-5.4}
& DR & 91.3 & 78.8 & 78.0 & 65.3 & 85.5 & 75.0 & 95.0 & 80.9 & 1.8 & \multirow{2}{*}{$-1.3$} \\
& EIA    & 94.0 & 78.0 & 84.7 & 49.3 & 86.5 & 74.0 & 91.0 & 79.6 & 3.5 & \\
\multirow{2}{*}{Gemini-3.1-Flash}
& DR & 89.3 & 73.6 & 63.3 & 62.7 & 72.5 & 71.0 & 91.0 & 74.2 & 0.9 & \multirow{2}{*}{$+1.4$} \\
& EIA    & 86.7 & 69.6 & 84.7 & 48.7 & 80.0 & 73.0 & 95.0 & 75.6 & 3.1 & \\
\multirow{2}{*}{Gemini-3.1-Pro}
& DR & 95.3 & 81.2 & 72.7 & 70.7 & 87.5 & 88.0 & 94.0 & 83.5 & 6.0 & \multirow{2}{*}{$+1.7$} \\
& EIA    & 94.0 & 84.4 & 92.0 & 64.0 & 88.5 & 86.0 & 88.0 & 85.2 & 3.4 & \\
\multirow{2}{*}{Claude-Haiku-4.5}
& DR & 90.0 & 73.2 & 72.7 & 64.0 & 82.0 & 77.0 & 79.0 & 76.6 & 2.5 & \multirow{2}{*}{$-0.4$} \\
& EIA    & 88.0 & 73.6 & 85.3 & 48.0 & 81.0 & 73.0 & 87.0 & 76.2 & 3.1 & \\
\multirow{2}{*}{Claude-Sonnet-4.6}
& DR & 90.7 & 73.2 & 61.3 & 68.7 & 84.0 & 77.0 & 85.0 & 76.7 & 1.2 & \multirow{2}{*}{$+7.4$} \\
& EIA    & 93.3 & 81.2 & 91.3 & 68.7 & 85.0 & 80.0 & 92.0 & 84.1 & 3.4 & \\
\bottomrule
\end{tabular}
}
\end{table}

\subsection{Experimental Setup}
We evaluate 13 representative LLMs on SMH-Bench under two settings. The first is Direct Reasoning (DR), where the model receives a global summary of the home state and the user query, then directly outputs the final prediction, such as structured actions, automation triggers, or natural-language responses. Predicted actions are executed in HomeEnv for outcome verification.
The second setting is the Environment-Interactive Agent (EIA), which follows the ReAct paradigm and interacts with HomeEnv through iterative observations and actions. 
Unlike Direct Reasoning, the agent starts with partial observability, such as basic room and device lists, and must use tools to query attributes, retrieve services, and execute operations. 
This setting better reflects real-world smart-home deployment. 
Implementation details for both settings are provided in Appendix~\ref{app:direct_setup}.

\begin{figure}[!t]
  \centering
  \begin{minipage}{0.48\textwidth}
    \centering
    \includegraphics[width=\textwidth]{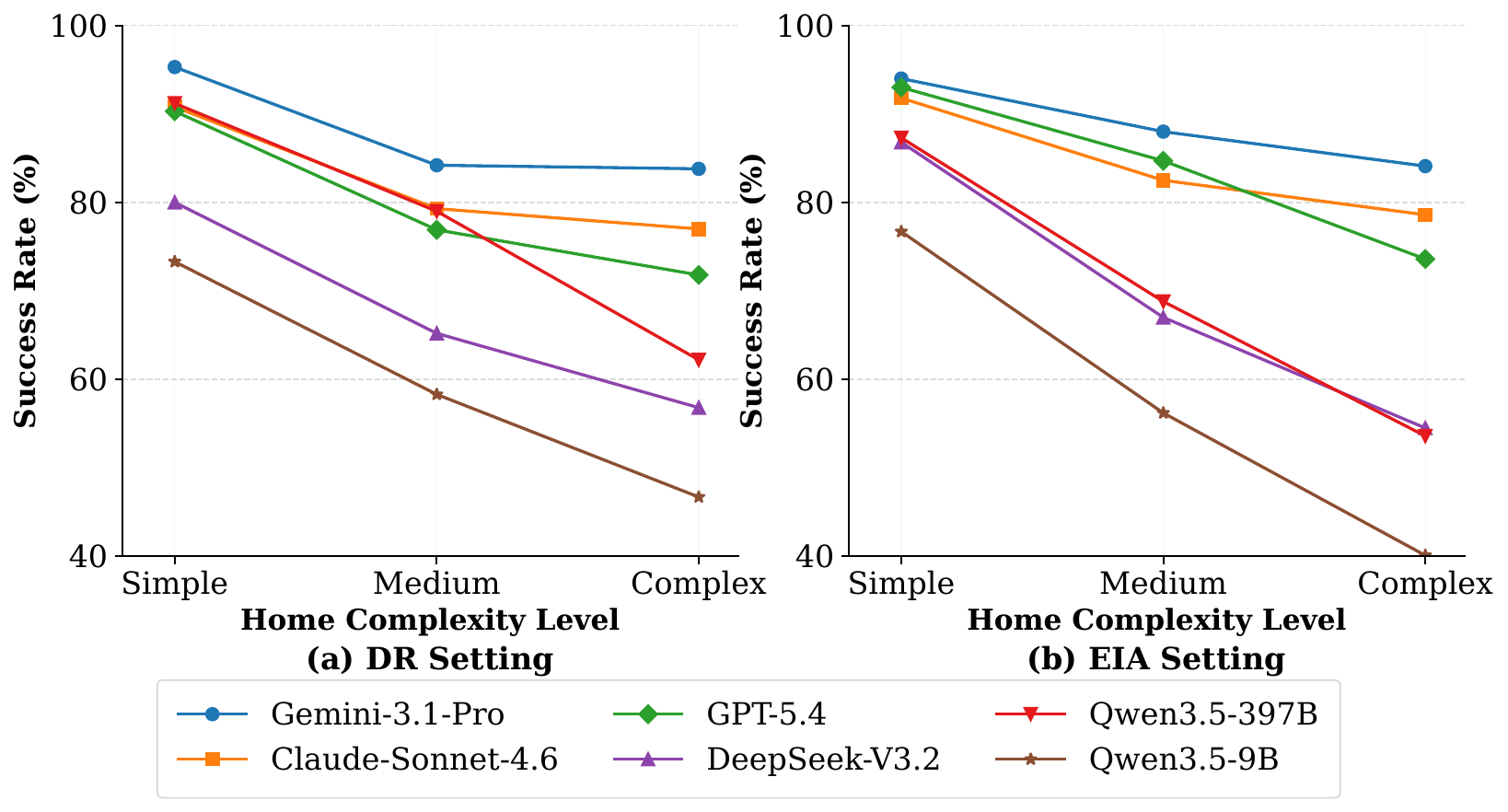}
    \caption{Pass rates of six models across home complexity levels in DR and EIA settings.}
    \label{fig:complexity}
  \end{minipage}
  \hfill
  \begin{minipage}{0.48\textwidth}
    \centering
    \includegraphics[width=\textwidth]{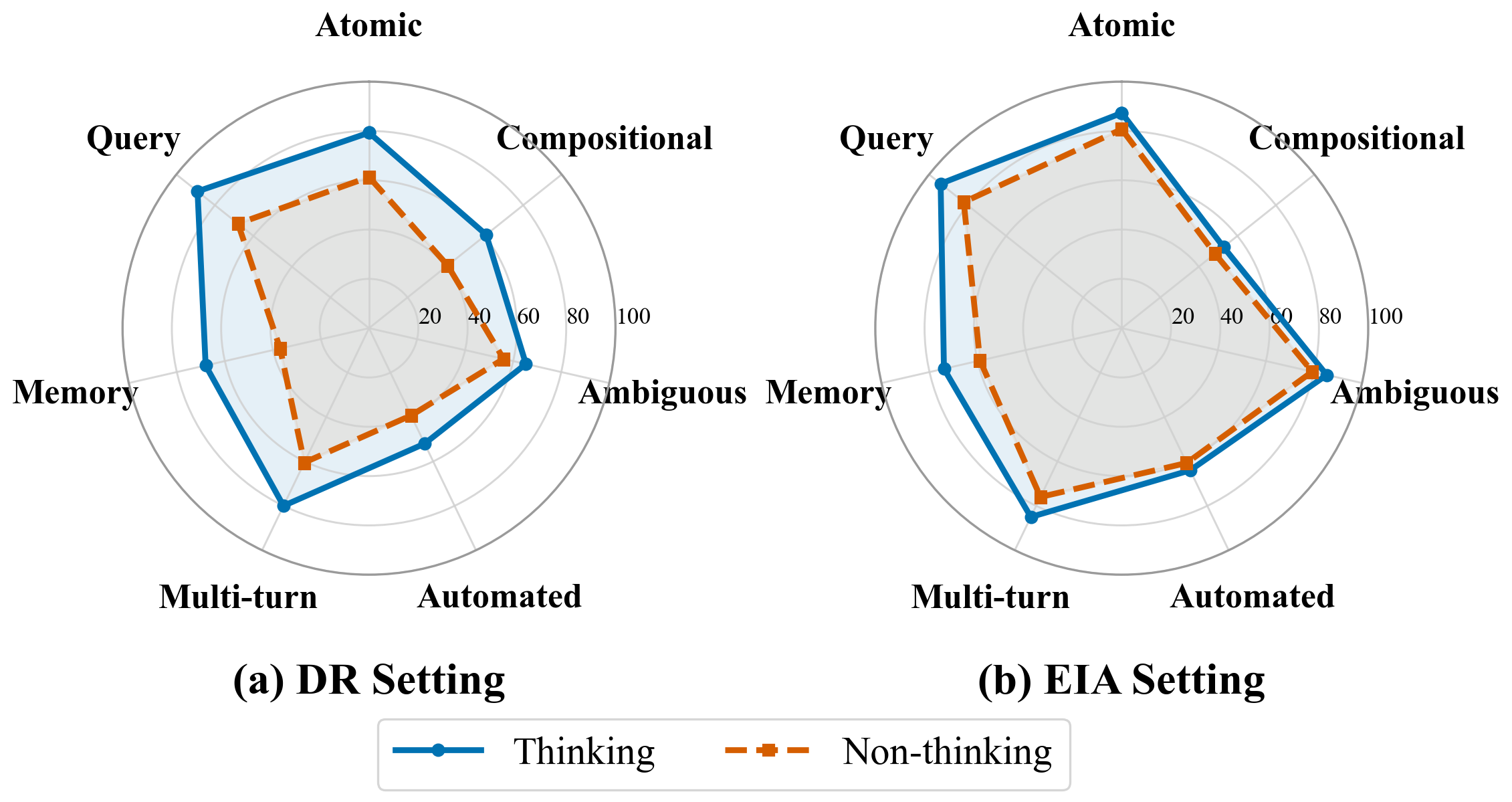}
    \caption{Comparison of DeepSeek-V3.2 Thinking and Non-thinking modes across task categories under the DR and EIA settings.}
    \label{fig:deepseek_radar}
  \end{minipage}
\end{figure}

\subsection{Main Results}

\paragraph{Overall Performance}
Table~\ref{tab:main_results} reports the main results of all evaluated models under the DR and EIA settings across seven capability dimensions. The strongest frontier model, Gemini-3.1-Pro, achieves an average success rate of 85.2\% under EIA and 83.5\% under DR. Among open-weight models, DeepSeek-V3.2 reaches 75.5\% under EIA and 69.5\% under DR, placing it on par with frontier compact variants.

\textbf{Capability-wise Insights.}
Across capability dimensions, model differences are most visible once tasks move beyond direct device access. TC1 and TC7 represent the most approachable categories; for instance, the leading Gemini-3.1-Pro achieves 94.0\% and 88.0\% under EIA respectively, and even our smallest evaluated model, Qwen3.5-4B, reaches a competitive 69.3\% and 76.0\%. However, the performance gap widens drastically on tasks requiring more complex reasoning. For example, DeepSeek-V3.2's EIA performance drops from 87.3\% on TC1 to merely 52.8\% on TC2. Furthermore, TC4 emerges as the most severe bottleneck across all models. Even the top-performing Gemini-3.1-Pro only manages 64.0\% under EIA and 70.7\% under DR on TC4, while our smallest evaluated model Qwen3.5-4B plummets to 21.3\% under EIA. These results highlight the multidimensional difficulty of SMH-Bench, validating its hierarchical design, which rigorously separates atomic device operation from high-level spatial, temporal, and compositional reasoning.

\textbf{Execution Mode Comparison.}
The EIA setting broadly benefits open-source and mid-tier models by enabling iterative exploration and real-time state querying. For instance, Qwen3.5-9B and GLM-5 exhibit massive average improvements of +15.1\% and +10.9\% under EIA, respectively. EIA is particularly critical for TC3, where models can iteratively clarify uncertainty; notably, Claude-Sonnet-4.6 jumps from 61.3\% under DR to 91.3\% under EIA in this category. Conversely, DR proves highly competitive for certain leading models with robust long-context integration. GPT-5.4 actually performs 1.3\% better overall under DR, particularly excelling in complex planning tasks like TC4, scoring 65.3\% under DR versus 49.3\% under EIA. This contrast demonstrates that while EIA strengthens interactive grounding, it introduces multi-step cascading errors that DR circumvents by relying entirely on the model's single-pass global reasoning.

\subsection{Further Analysis}






\paragraph{Impact of Home Complexity.}
LLMs become less reliable as home environments grow more complex. As shown in Figure~\ref{fig:complexity}, all representative models exhibit a clear performance drop from simple to complex homes in both settings. In the DR setting, Gemini-3.1-Pro decreases from 95.3\% to 83.8\%, while Qwen3.5-9B drops more sharply from 73.3\% to 46.7\%. A similar trend appears in the EIA setting. These results suggest that home complexity mainly hurts models by expanding the space of candidate rooms, devices, states, and constraints. Tool access helps retrieve states and locate devices, but it does not solve the higher-level problem of deciding what to query, what to ignore, and when to stop. The gap therefore shows that current agents still struggle to scale to realistic homes, where execution requires robust grounding, state filtering, and constraint satisfaction.



\paragraph{When EIA Compensates for Thinking.}
As shown in Figure~\ref{fig:deepseek_radar}, removing thinking hurts DeepSeek-V3.2 much more in DR than in EIA, with the overall drop increasing from 7.3 points in EIA to 18.3 points in DR. The non-thinking variant remains relatively competitive in EIA on atomic control and device query, suggesting that tool calls can externalize state retrieval and action execution. In contrast, the largest drops in DR appear on memory, query, composite, and multi-turn tasks, where the model must internally select relevant evidence, maintain user preferences, and integrate information across steps. This contrast shows that tool interaction can compensate for weak reasoning when tasks depend mainly on observable states and explicit actions, but thinking remains crucial for latent preferences, multi-step coordination, and deciding what information to query or ignore.

\paragraph{Subcategory-level Paradigm Preferences.}
The fine-grained subcategory results show a clear division between tasks that benefit from interaction and tasks that favor one-shot reasoning. DR is consistently stronger on atomic and composite control, as well as memory and multi-turn subcategories, where the relevant evidence is already present in the prompt and the model mainly needs to parse instructions, integrate context, and generate a valid structured action. EIA is more suitable for ambiguous intent and device-query tasks, where tool calls can resolve missing slots, inspect current states, and ground references in the environment. Several subcategories show only marginal gaps, including automation, multi-turn variants, and simple query or control cases, suggesting that the limiting factor is not the execution paradigm itself but the underlying ability to synthesize conditions, track dialogue updates, or map natural language to device constraints. Complete results for all 22 subcategories are reported in Appendix~\ref{sec:appendix_examples}.

\begin{table}[t]
\centering
\small
\caption{Error taxonomy. Detailed descriptions and cases are provided in Appendix~\ref{app:error-distribution}.}
\label{tab:error-taxonomy}
\begin{tabular}{p{0.32\linewidth} p{0.58\linewidth}}
\toprule
\textbf{Error Type} & \textbf{Scope} \\
\midrule
Structured Output (SO)
& Missing or malformed structured fields, including labels, automation conditions, query responses, or empty outputs. \\

\midrule
Mode Selection (MS)
& Incorrectly choosing between execution and clarification. \\

\midrule
Automation Judging (AJ)
& Incorrect or failed validation of automation conditions. \\

\midrule
Action/Tool Validity (AV/TV)
& Invalid tool calls, including missing arguments, invalid \texttt{None} values, wrong names, type errors, or other schema violations. \\

\midrule
Interaction Efficiency (IE)
& Inefficient interaction behavior, especially excessive tool use. \\

\midrule
Format/Runtime Failure (FRF)
& Unparseable JSON output that prevents the evaluator from extracting executable actions. \\
\bottomrule
\end{tabular}
\end{table}

\begin{figure}[t]
  \centering
  \includegraphics[width=0.6\textwidth]{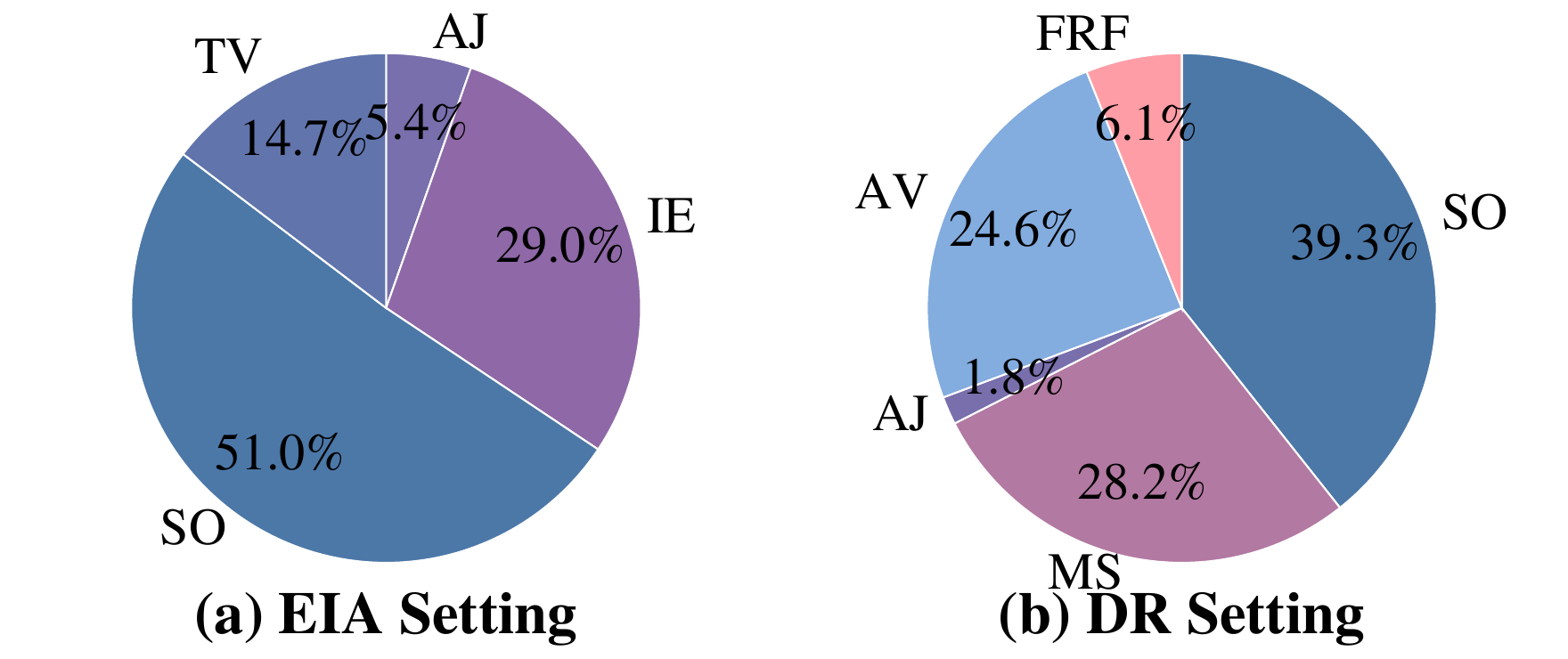}
  \caption{Error type distributions of DeepSeek-V3.2 on DR and EIA.}
  \label{fig:error-type}
\end{figure}

\subsection{Error Analysis}
We analyzed the failure modes of existing models when converting smart-home intents into structured actions.
As summarized in Table~\ref{tab:error-taxonomy}, these models exhibit diverse errors when generating executable actions. These errors span structured output (SO), mode selection (MS), automation judgment (AJ), action/tool validity (AV/TV), interaction efficiency (IE), and format/runtime robustness (FRF).
Based on these six error dimensions, we analyze DeepSeek-V3.2 failure cases under the EIA and DR settings.
As shown in Figure~\ref{fig:error-type}, SO remains the predominant error source in both settings, indicating models still struggle to follow the specified schema in constrained tasks. In the EIA setting, IE errors are notably prominent, likely because models tend to invoke redundant tools when processing ambiguous instructions and lose parameter boundaries during multi-turn interactions. By contrast, the DR setting primarily features MS and AV errors. This pattern emerges because models lacking interactive feedback tend to execute actions directly with insufficient information.
Overall, these results show that SMH-Bench evaluates more than semantic understanding, as success also requires state grounding, appropriate clarification policies, schema adherence, condition synthesis, and efficient tool interaction. Appendix~\ref{app:error-distribution} provides full distributions and case studies.


%% file: sections/5.conclusion.tex
\section{Conclusion}

We introduced SMH-Bench, a simulator-grounded benchmark for evaluating LLM-driven agents in realistic smart-home settings. Built on HomeEnv, SMH-Bench moves beyond static instruction-to-API matching by verifying agents through executable state transitions. It contains 1,100 human-audited tasks across seven capability categories and 22 fine-grained subcategories, spanning homes from small apartments to nested multi-room environments with up to 135 devices. Our evaluation of 13 representative LLMs under both DR and EIA settings shows that current models handle explicit control and simple queries relatively well, but remain unreliable on automation, ambiguity resolution, personalized memory, and complex-home grounding. These results suggest that future smart-home agents need stronger state grounding, clarification policies, preference-aware reasoning, and schema-reliable tool use. We hope SMH-Bench will support future progress toward more reliable, context-aware, and deployable smart-home agents.


%% file: appendix/mha_state_space1.tex
\section{Smart-Home State Space Schema in HomeEnv}
\label{app:HomeEnv_state_space}

This appendix specifies the executable HomeEnv state schema used by SMH-Bench. HomeEnv represents a home through rooms and devices; devices expose typed attributes and validated services, while multi-component devices use component namespaces to disambiguate attribute and service locators. This schema is the state representation used for action replay, state-difference extraction, and rule-based verification in the benchmark.

\subsection{Home-Level State}

A HomeEnv state is stored as a structured home object with two top-level entries: \texttt{rooms} and \texttt{devices}. Rooms provide the spatial layout of the home, while devices provide executable entities assigned to rooms. Each device contains either device-level or component-level attributes and services, which together support grounding, action execution, state-difference extraction, and rule-based verification.

\begin{table}[!htbp]
\centering
\caption{Top-level objects in the HomeEnv smart-home state space.}
\label{tab:HomeEnv_top_level_schema}
\small
\begin{tabular}{@{}p{0.13\linewidth}p{0.34\linewidth}p{0.45\linewidth}@{}}
\toprule
Object & Role & Relation \\
\midrule
Home & Executable environment state. & Contains \texttt{rooms} and \texttt{devices}. \\
Room & Spatial unit used for grounding. & Identified by \texttt{id}; referenced by device metadata. \\
Device & Controllable or observable entity. & Assigned to exactly one room through \texttt{userdata.room}. \\
Attribute & Typed state variable. & Belongs to a device or component. \\
Service & Valid state transition. & Belongs to a device or component and updates attributes after validation. \\
Component & Functional namespace for multi-function devices. & Groups attributes and services under paths such as \texttt{light.*}. \\
\bottomrule
\end{tabular}
\end{table}

\FloatBarrier
\subsection{Room Schema}

Rooms define the spatial structure of the simulated home. Each room has a unique ID, a semantic type, a user-facing name, and a floor index. HomeEnv also supports nested room structures through the optional \texttt{parent} field, allowing environments to represent spaces such as a bathroom inside a bedroom or a balcony attached to a bedroom.

\begin{table}[!htbp]
\centering
\caption{Room schema.}
\label{tab:HomeEnv_room_schema}
\small
\begin{tabular}{@{}p{0.18\linewidth}p{0.18\linewidth}p{0.56\linewidth}@{}}
\toprule
Field & Type & Description \\
\midrule
\texttt{id} & string & Unique room ID. \\
\texttt{type} & string & Semantic room type. \\
\texttt{name} & string & User-facing room name used in instructions and observations. \\
\texttt{floor} & integer & Floor index. \\
\texttt{parent} & string, optional & Parent room ID for nested spaces. \\
\bottomrule
\end{tabular}
\end{table}

\FloatBarrier
\subsection{Device Schema}

Devices connect spatial grounding with executable control. Each device is assigned to one room through \texttt{userdata.room}, exposes dynamic state through attributes, and exposes valid state transitions through services. Devices with multiple functional parts, such as fan-lights, use \texttt{components} instead of a single flat attribute-service namespace.

\begin{table}[!htbp]
\centering
\caption{Device schema.}
\label{tab:HomeEnv_device_schema}
\small
\begin{tabular}{@{}p{0.18\linewidth}p{0.16\linewidth}p{0.58\linewidth}@{}}
\toprule
Field & Type & Description \\
\midrule
\texttt{name} & string & Human-readable device name. \\
\texttt{description} & string & Natural-language device description. \\
\texttt{attributes} & list & Device-level state variables. Empty when all functions are component-level. \\
\texttt{services} & list & Device-level executable operations. Empty when all functions are component-level. \\
\texttt{components} & list & Optional component namespaces, each with attributes and services. \\
\texttt{userdata} & object & Static metadata for identity, retrieval, and room assignment. \\
\bottomrule
\end{tabular}
\end{table}

\begin{table}[!htbp]
\centering
\caption{Device metadata fields in \texttt{userdata}.}
\label{tab:HomeEnv_device_metadata_schema}
\small
\begin{tabular}{@{}p{0.19\linewidth}p{0.18\linewidth}p{0.55\linewidth}@{}}
\toprule
Field & Type & Description \\
\midrule
\texttt{did} & string & Unique device instance ID. \\
\texttt{spid} & string & Product specification ID shared by devices with the same capability schema. \\
\texttt{category} & string & Coarse device category. \\
\texttt{subcategory} & string & Fine-grained device type. \\
\texttt{tags} & list[string] & Semantic tags used for retrieval and filtering. \\
\texttt{room} & string & ID of the room where the device is located. \\
\bottomrule
\end{tabular}
\end{table}

Table~\ref{tab:HomeEnv_device_categories} summarizes representative device categories covered by HomeEnv.

\begin{table}[!htbp]
\centering
\caption{Representative device categories.}
\label{tab:HomeEnv_device_categories}
\small
\begin{tabular}{@{}p{0.26\linewidth}p{0.66\linewidth}@{}}
\toprule
Category & Representative subcategories \\
\midrule
\texttt{light} & light, color light, color-temperature light, light strip \\
\texttt{sensor} & climate sensor, human sensor \\
\texttt{fan\_light} & fan light \\
\texttt{media} & smart screen, universal speaker \\
\texttt{ac} & AC gateway, air cube, floor AC \\
\texttt{vacuum} & vacuum cleaner, robot cleaner \\
\texttt{water\_heater} & electric water heater \\
\texttt{lock} & smart lock \\
\texttt{cover} & curtain, blind \\
\texttt{fan} & smart fan \\
\texttt{washing\_machine} & compound washer \\
\texttt{oven} & built-in oven \\
\bottomrule
\end{tabular}
\end{table}

\FloatBarrier
\subsection{Attribute Schema}

Attributes are the basic state variables in HomeEnv. Each attribute defines a state name, type, optional value constraints, and current value. These constraints are used both during service execution and task verification. HomeEnv supports scalar types such as \texttt{str}, \texttt{bool}, \texttt{int}, \texttt{float}, \texttt{hex}, and \texttt{enum}, as well as compound types such as \texttt{list}, \texttt{tuple}, and \texttt{dict}.

\begin{table}[!htbp]
\centering
\caption{Attribute schema.}
\label{tab:HomeEnv_attribute_schema}
\small
\begin{tabular}{@{}p{0.18\linewidth}p{0.18\linewidth}p{0.56\linewidth}@{}}
\toprule
Field & Type & Description \\
\midrule
\texttt{name} & string & Attribute name. \\
\texttt{description} & string & Natural-language description of the state variable. \\
\texttt{type} & string & Data type, such as \texttt{str}, \texttt{int}, \texttt{float}, \texttt{bool}, or \texttt{tuple}. \\
\texttt{value} & typed value & Current attribute value. \\
\texttt{range} & list, optional & Numeric lower and upper bounds. \\
\texttt{options} & list, optional & Finite set of allowed values. \\
\texttt{items} & object, optional & Element schema for compound values. \\
\texttt{precision} & number, optional & Numeric precision constraint. \\
\texttt{unit} & string, optional & Unit, such as percentage, Celsius, Kelvin, or degree. \\
\bottomrule
\end{tabular}
\end{table}

\begin{table}[!htbp]
\centering
\caption{Representative attributes.}
\label{tab:HomeEnv_representative_attributes}
\small
\begin{tabular}{@{}p{0.14\linewidth}p{0.27\linewidth}p{0.28\linewidth}p{0.21\linewidth}@{}}
\toprule
Device category & Attribute & Type / constraint & Meaning \\
\midrule
General device & \texttt{state} & \texttt{str}: \texttt{on}, \texttt{off} & Power state. \\
General device & \texttt{battery\_level} & \texttt{int}, $[0,100]$ & Battery percentage. \\
Light & \texttt{brightness} & \texttt{int}, $[1,100]$ & Brightness percentage. \\
Light & \texttt{color\_temperature} & \texttt{int}, $[2700,6500]$ & Color temperature. \\
AC & \texttt{target\_temperature} & \texttt{float}, device-specific range & Target temperature. \\
AC & \texttt{ac\_mode} & \texttt{str}: heat, cool, auto, dry, fan\_only & AC mode. \\
Fan & \texttt{speed\_level} & \texttt{str}: low, medium, high & Fan speed level. \\
Sensor & \texttt{current\_temperature} & \texttt{float} & Measured temperature. \\
Sensor & \texttt{current\_humidity} & \texttt{float} & Measured humidity. \\
Sensor & \texttt{human\_detected} & \texttt{str}: yes, no & Human-presence status. \\
Curtain & \texttt{position} & \texttt{int}, $[0,100]$ & Opening percentage. \\
Speaker & \texttt{volume\_level} & \texttt{int}, $[0,100]$ & Playback volume. \\
Vacuum & \texttt{suction} & \texttt{str}: low, medium, high, auto & Suction level. \\
Lock & \texttt{alarm\_volume} & \texttt{int}, $[0,100]$ & Alarm volume. \\
\bottomrule
\end{tabular}
\end{table}

\FloatBarrier
\subsection{Service Schema}

Services define executable operations that agents may call to modify the home state. A service contains a name, description, optional arguments, and an execution rule. Service arguments reuse the attribute type system, so HomeEnv can validate argument names, types, ranges, precision, and enum options before applying the state transition rule. Invalid service calls are rejected and do not mutate the home state.

\begin{table}[!htbp]
\centering
\caption{Service schema.}
\label{tab:HomeEnv_service_schema}
\small
\begin{tabular}{@{}p{0.18\linewidth}p{0.18\linewidth}p{0.56\linewidth}@{}}
\toprule
Field & Type & Description \\
\midrule
\texttt{name} & string & Service name. \\
\texttt{description} & string & Natural-language description of the operation. \\
\texttt{arguments} & list & Required or optional parameters, using the same constraints as attributes. \\
\texttt{code} & string & State transition rule executed after validation succeeds. \\
\bottomrule
\end{tabular}
\end{table}

\begin{table}[!htbp]
\centering
\caption{Representative services.}
\label{tab:HomeEnv_representative_services}
\small
\begin{tabular}{@{}p{0.14\linewidth}p{0.30\linewidth}p{0.24\linewidth}p{0.22\linewidth}@{}}
\toprule
Device category & Service & Arguments & State update \\
\midrule
General device & \texttt{turn\_on} & none & \texttt{state = "on"} \\
General device & \texttt{turn\_off} & none & \texttt{state = "off"} \\
Light & \texttt{set\_brightness} & \texttt{brightness} & Updates brightness. \\
Light & \texttt{set\_color\_temperature} & \texttt{color\_temperature} & Updates color temperature. \\
AC & \texttt{set\_ac\_mode} & \texttt{ac\_mode} & Updates AC mode. \\
AC & \texttt{set\_target\_temperature} & \texttt{target\_temperature} & Updates target temperature. \\
Fan & \texttt{set\_speed\_level} & \texttt{speed\_level} & Updates fan speed. \\
Curtain & \texttt{open}, \texttt{close} & none & Opens or closes the curtain. \\
Curtain & \texttt{set\_position} & \texttt{position} & Updates curtain position. \\
Speaker & \texttt{play}, \texttt{pause}, \texttt{stop} & metadata or none & Updates playback state. \\
Vacuum & \texttt{start, pause}, \texttt{return\_to\_base} & none & Updates the cleaning state of devices. \\
Lock & \texttt{lock}, \texttt{open} & none & Locks or unlocks the device. \\
\bottomrule
\end{tabular}
\end{table}

\FloatBarrier
\subsection{Multi-Component Devices}

Multi-component devices use component namespaces to prevent ambiguity between functions hosted by the same physical device. For example, a fan-light device contains a \texttt{light} component and a \texttt{fan} component. Their attributes and services are addressed through namespaced paths, such as \texttt{light.brightness} or \texttt{fan.set\_speed\_level}. This convention allows the evaluator to distinguish whether an action targets the light function or the fan function of the same device ID.

\begin{table}[!htbp]
\centering
\caption{Example namespace schema for a fan-light device.}
\label{tab:HomeEnv_fan_light_schema}
\small
\begin{tabular}{@{}p{0.12\linewidth}p{0.45\linewidth}p{0.35\linewidth}@{}}
\toprule
Component & Attribute namespace & Service locator \\
\midrule
\texttt{light} & \texttt{light.state, light.brightness}, \texttt{light.color\_temperature} & \texttt{light.turn\_on}, \texttt{light.set\_brightness} \\
\texttt{fan} & \texttt{fan.state, fan.speed\_level}, \texttt{fan.preset\_mode} & \texttt{fan.turn\_on}, \texttt{fan.set\_speed\_level} \\
\bottomrule
\end{tabular}
\end{table}

\FloatBarrier
\subsection{Home-Level Example}

The following valid JSON fragment illustrates how HomeEnv connects rooms, device attributes, service schemas, component namespaces, and room assignments within one executable home state. The example omits unrelated rooms and devices for brevity.

\begin{schemaformatbox}{Home-level Example}
{
  "name": "home",
  "rooms": [
    {"id": "10", "type": "dining", "name": "dining room", "floor": 1},
    {"id": "25", "type": "study", "name": "study room", "floor": 2},
    {"id": "12", "type": "bathroom", "name": "master-bedroom bathroom", "floor": 1, "parent": "21"}
  ],
  "devices": [
    {
      "name": "floor-standing AC",
      "description": "floor-standing AC",
      "attributes": [
        {
          "name": "state",
          "description": "current power state of the device",
          "type": "str",
          "options": ["on", "off"],
          "value": "off"
        },
        {
          "name": "target_temperature",
          "description": "target temperature",
          "type": "float",
          "range": [18, 30],
          "precision": 1,
          "value": 18,
          "unit": "celsius"
        },
        {
          "name": "ac_mode",
          "description": "current AC mode",
          "type": "str",
          "options": ["heat", "cool", "auto", "dry", "fan_only"],
          "value": "heat"
        }
      ],
      "services": [
        {
          "name": "turn_on",
          "description": "turn the device on",
          "arguments": [],
          "code": "self.state = \"on\""
        },
        {
          "name": "set_target_temperature",
          "description": "set the target temperature",
          "arguments": [
            {
              "name": "target_temperature",
              "type": "float",
              "range": [18, 30],
              "precision": 1,
              "unit": "celsius"
            }
          ],
          "code": "self.target_temperature = target_temperature"
        }
      ],
      "components": [],
      "userdata": {
        "did": "1693",
        "spid": "100011197",
        "category": "ac",
        "subcategory": "floor_ac",
        "tags": ["ac"],
        "room": "10"
      }
    },
    {
      "name": "fan light",
      "description": "fan light",
      "attributes": [],
      "services": [],
      "components": [
        {
          "name": "light",
          "attributes": [
            {
              "name": "state",
              "description": "current power state of the light component",
              "type": "str",
              "options": ["on", "off"],
              "value": "on"
            },
            {
              "name": "brightness",
              "description": "current light brightness",
              "type": "int",
              "range": [1, 100],
              "value": 58,
              "unit": "percentage"
            }
          ],
          "services": [
            {
              "name": "turn_on",
              "description": "turn the light component on",
              "arguments": [],
              "code": "self.light.state = \"on\""
            },
            {
              "name": "set_brightness",
              "description": "set the light brightness",
              "arguments": [
                {
                  "name": "brightness",
                  "description": "target light brightness",
                  "type": "int",
                  "range": [1, 100],
                  "unit": "percentage"
                }
              ],
              "code": "self.light.brightness = brightness"
            }
          ]
        },
        {
          "name": "fan",
          "attributes": [
            {
              "name": "state",
              "description": "current power state of the fan component",
              "type": "str",
              "options": ["on", "off"],
              "value": "off"
            },
            {
              "name": "speed_level",
              "description": "current fan speed level",
              "type": "str",
              "options": ["low", "medium", "high"],
              "value": "medium"
            }
          ],
          "services": [
            {
              "name": "turn_on",
              "description": "turn the fan component on",
              "arguments": [],
              "code": "self.fan.state = \"on\""
            },
            {
              "name": "set_speed_level",
              "description": "set the fan speed level",
              "arguments": [
                {
                  "name": "speed_level",
                  "description": "target fan speed level",
                  "type": "str",
                  "options": ["low", "medium", "high"]
                }
              ],
              "code": "self.fan.speed_level = speed_level"
            }
          ]
        }
      ],
      "userdata": {
        "did": "3460",
        "spid": "20000010",
        "category": "fan_light",
        "subcategory": "fan_light",
        "tags": ["fan", "light"],
        "room": "25"
      }
    },
    {
      "...": "additional devices omitted"
    }
  ]
}
\end{schemaformatbox}

\FloatBarrier

%% file: appendix/task_instruction_generation_rules_backup4.tex
\section{Generation Rules and Examples for Task and Instruction Construction}
\label{app:task_instruction_generation}

This appendix details how SMH-Bench constructs user instructions and executable references from a home environment. In the implementation, Task and Instruction Generation is a \emph{template-guided generation and grounding} procedure. It does not require a universal intermediate representation that stores a complete task specification before the user utterance is written. Instead, a generated sample is grounded by the combination of (i) the task-specific user-generation template, (ii) the generated user instruction, (iii) the reference assistant JSON, and (iv) replay-derived labels and checks.

\subsection{Template-Guided Generation and Grounding}

The generation flow is organized as four sequential stages. The user-facing instruction is produced first under a task-template constraint; the executable target is then recovered through a reference assistant response and simulator replay. This design keeps the language natural while ensuring that every benchmark sample is grounded in devices, services, and states that actually exist in the home.

\textbf{Stage 1: Home-state serialization.} The generation flow first serializes the HomeEnv instance into a textual environment containing rooms, devices, attributes, supported services, scenes, the entry device, and optional dialogue or memory context. This serialized view is the only state evidence shown to later generation calls, so it constrains the task to valid device capabilities and prevents references to nonexistent rooms or unsupported operations.

\textbf{Stage 2: User instruction generation.} Given the serialized home, chat history, and one sampled user-generation template, the first LLM call writes the user side of the task. Its output contains an \texttt{analysis} field and a \texttt{user\_query}; templates that deliberately create ambiguity may additionally preserve an \texttt{intended\_query}. The template fixes the task family, target-selection rule, and surface style, while the LLM realizes these constraints as a realistic household utterance.

\textbf{Stage 3: Reference assistant generation.} The generated \texttt{user\_query} is then paired with the same home state and dialogue history in a second LLM call. The reference assistant returns JSON with a \texttt{mode}, a natural-language \texttt{response}, and one of executable \texttt{actions}, \texttt{automations}, or a clarification. This step grounds the utterance back to concrete device identifiers, service locators, argument values, automation conditions, query answers, or clarification behavior.

\textbf{Stage 4: Replay and label extraction.} Finally, the reference trajectory is replayed against the executable HomeEnv state. Replay records the state changes caused by the reference actions and converts them into deterministic evaluation targets, including \texttt{ground\_truth.labels}, \texttt{ground\_truth.conditions}, and \texttt{critic}. As a result, different surface phrasings can be accepted during generation, but evaluation is performed against stable device-level outcomes.

\begin{promptformatbox}{Generation example}
## Stage 1: Home-state evidence
Example:
The living room contains two controllable lights.
Living-room main light:
- did = 1001
- attribute = light.brightness: 85
- service = light.set_brightness(brightness: integer, range 1..100)

Living-room downlight:
- did = 1002
- attribute = light.brightness: 90
- service = light.set_brightness(brightness: integer, range 1..100)

The home also contains one bedroom lamp:
Bedroom lamp:
- did = 2001
- attribute = light.brightness: 35
- service = light.set_brightness(brightness: integer, range 1..100)

## Stage 2: User instruction generation
Example output:
{
  "analysis": "The living room has two lights with brightness above 80, while the bedroom lamp is only 35 and is outside the requested room. The instruction should dim only the living-room lights satisfying the threshold.",
  "user_query": "Set the living-room lights above 80
}

## Stage 3: Reference assistant generation
Example output:
{
  "mode": "execution",
  "response": "The living-room lights above 80
  "actions": [
    {
      "did": "1001",
      "locator": "light.set_brightness",
      "arguments": {"brightness": 40}
    },
    {
      "did": "1002",
      "locator": "light.set_brightness",
      "arguments": {"brightness": 40}
    }
  ]
}

## Stage 4: Replay-derived labels
Example output:
{
  "ground_truth": {
    "labels": [
      {"did": "1001", "attribute": "light.brightness", "value": 40},
      {"did": "1002", "attribute": "light.brightness", "value": 40}
    ]
  },
  "critic": "device('1001').light.brightness == 40 and device('1002').light.brightness == 40"
}

\end{promptformatbox}

\FloatBarrier

\subsection{Template Rules and Constraints}

The first LLM call samples one task-specific template. The template determines which target pattern can be selected from the home state and how the user instruction should be expressed. Table~\ref{tab:user_template_rules} lists the complete template families used for Task and Instruction Generation.

\begingroup
\small
\setlength{\tabcolsep}{3pt}
\begin{longtable}{>{\raggedright\arraybackslash}p{0.20\linewidth}
                  >{\raggedright\arraybackslash}p{0.23\linewidth}
                  >{\raggedright\arraybackslash}p{0.29\linewidth}
                  >{\raggedright\arraybackslash}p{0.22\linewidth}}
\caption{Template families and generation rules.}
\label{tab:user_template_rules}\\
\toprule
Family & Templates & Grounding rule & Surface instruction rule \\
\midrule
\endfirsthead
\caption[]{Template families and generation rules (continued).}\\
\toprule
Family & Templates & Grounding rule & Surface instruction rule \\
\midrule
\endhead
\multicolumn{4}{r}{Continued on next page} \\
\midrule
\endfoot
\bottomrule
\endlastfoot

\texttt{TC1: Atomic}\newline
\texttt{Control} &
\texttt{Clear Commands};\newline
\texttt{Colloquial Requests};\newline
\texttt{Noisy Utterances} &
Select one controllable device and one supported service or attribute change. The target must be unique and directly executable. &
\texttt{clear} is explicit; \texttt{colloquial} uses everyday phrasing; \texttt{noisy} may include filler words, repetition, ASR-like artifacts, or immediate self-correction. \\
\midrule
\texttt{TC2: Compositional}\newline
\texttt{Control} &
\texttt{Explicit Multi-device};\newline
\texttt{Batch Operations};\newline
\texttt{State-dependent};\newline
\texttt{Room-dependent};\newline
\texttt{Top-N Selection} &
Select multiple operations, a device set, a state-filtered subset, a room-conditioned subset, or a ranked top-$k$ target. All selected devices must support the requested operation. &
Express the dependency in one household request, e.g., ``all lights'', ``lights above 80\%'', ``rooms where the AC is on'', or ``the two fastest fans''. \\
\midrule
\texttt{TC3: Ambiguous}\newline
\texttt{Intent} &
\texttt{State-level Ambiguity};\newline
\texttt{Scene-level Ambiguity};\newline
\texttt{Slot-missing} &
Choose a plausible target but make the initial utterance under-specified, subjective, or scene-like. Missing slots may include room, device, mode, value, duration, or time. &
The initial query should require clarification or context-based inference. The hidden target is preserved through \texttt{intended\_query} or the reference clarification trajectory. \\
\midrule
\texttt{TC4: Automated}\newline
\texttt{Task}\newline
\texttt{Scheduling} &
\texttt{Time-specific};\newline
\texttt{State-driven};\newline
\texttt{Time-and-state} &
Choose supported actions and attach a delay, scheduled time, state trigger, or combined time-state condition. &
Use natural scheduling language such as ``in two hours'', ``every morning at 7'', ``when the window opens'', or ``at night if the humidity is high''. \\
\midrule
\texttt{TC5: Context-Aware}\newline
\texttt{Multi-turn}\newline
\texttt{Interaction} &
\texttt{Context Continuation};\newline
\texttt{Topic Shifts};\newline
\texttt{Coreference Resolution};\newline
\texttt{User Corrections} &
The current target depends on previous turns. The generator may inherit a device, override a previous operation, resolve a pronoun, or switch topics while preserving context. &
Use natural dialogue forms such as ``also do that'', ``no, make it 26'', ``turn that one off'', or a topic-shifted follow-up. \\
\midrule
\texttt{TC6: Personalized}\newline
\texttt{Memory} &
\texttt{Short-term Memory};\newline
\texttt{Long-term Memory} &
Use recent interaction facts or stored user preferences to fill omitted action details. The task should not be solvable in the intended way without the provided memory. &
Allow compact preference-triggering instructions such as ``sleep time'', ``as usual'', or ``set it how I like it''. \\
\midrule
\texttt{TC7: Environment-}\newline
\texttt{Grounded Query} &
\texttt{Status Inquiries};\newline
\texttt{Quantity Queries} &
Compute the answer from the home state, device attributes, or available tools without emitting device-control actions. &
Ask a natural question about counts, locations, current status, extrema, comparisons, or device availability. \\
\end{longtable}
\endgroup

\FloatBarrier

%% file: appendix/task_taxonomy.tex
\section{Detailed Task Taxonomy}
\label{sec:appendix_examples}

This appendix expands the task taxonomy. SMH-Bench contains 22 fine-grained subcategories under seven primary task categories. Each subcategory contains 50 evaluation instances in the released diverse subset, with 25 simple-home instances, 15 medium-home instances, and 10 complex-home instances. Table~\ref{tab:taxonomy_all} consolidates the full taxonomy into a single table.

\newcommand{\taxgroup}[2]{\multicolumn{3}{@{}l}{\small\textbf{#1. #2}}\\\midrule}
\newcommand{\taxcode}[1]{\par\noindent{\ttfamily\scriptsize\sloppy #1}}
\newcommand{\taxblock}[2]{\noindent\textbf{#1.} #2}
\newcommand{\taxcodeblock}[2]{\noindent\textbf{#1.}\taxcode{#2}}
\newcommand{\taxsub}[1]{\textbf{#1}}
\newcommand{\taxrowsep}{\specialrule{0.25pt}{2pt}{2pt}}
\newcommand{\evaltarget}[2]{\taxblock{Target}{#1}\par\vspace{5pt}\taxcodeblock{Critic}{#2}}
\newcommand{\evalauto}[3]{\taxblock{Target}{#1}\par\vspace{5pt}\taxcodeblock{Trigger}{#2}\par\vspace{5pt}\taxcodeblock{Critic}{#3}}
\newcommand{\evalquery}[2]{\taxblock{Target}{#1}\par\vspace{5pt}\taxblock{Expected answer}{#2}\par\vspace{5pt}\taxcodeblock{Critic}{N/A (non-mutating query)}}

\begingroup
\scriptsize
\setlength{\tabcolsep}{4pt}
\renewcommand{\arraystretch}{1.34}
\setlength{\extrarowheight}{2pt}
\begin{longtable}{>{\raggedright\arraybackslash}p{0.145\linewidth}
                  >{\raggedright\arraybackslash}p{0.375\linewidth}
                  >{\raggedright\arraybackslash}p{0.42\linewidth}}
\caption{Full task taxonomy with representative examples and critic-style evaluation targets.}
\label{tab:taxonomy_all}\\
\toprule
Subcategory & Representative instruction & Evaluation target \\
\midrule
\endfirsthead
\toprule
Subcategory & Representative instruction & Evaluation target \\
\midrule
\endhead
\multicolumn{3}{r}{Continued on next page} \\
\midrule
\endfoot
\bottomrule
\endlastfoot

\taxgroup{TC1: Atomic Control}
\taxsub{Clear Commands} & ``Set the water heater in the west kitchen to a target water temperature of 45 degrees.'' & \evaltarget{Set the west-kitchen water heater target water temperature to 45.}{device(9470).\allowbreak target\_\allowbreak water\_\allowbreak temperature == 45} \\
\taxrowsep

\taxsub{Colloquial Requests} & ``Put the main-bedroom fan into blowing mode.'' & \evaltarget{Turn on the main-bedroom fan and set normal mode.}{device(4267).\allowbreak state == on and device(4267).\allowbreak preset\_\allowbreak mode == normal} \\
\taxrowsep

\taxsub{Noisy Utterances} & ``Set the bedroom temperature-humidity sensor, no, wait, the living-room sensor, alias to living-room sensor.'' & \evaltarget{Apply the corrected target, not the discarded one.}{device(3960).\allowbreak alias == living-room sensor} \\
\taxrowsep

\taxgroup{TC2: Compositional Control}
\taxsub{Explicit Multi-device} & ``Resume music on the spare-bedroom smart screen and set its volume to 40; open and relock the stuck door lock; set the dining-room colored light to 30\% brightness; turn on the kitchen fan in natural mode.'' & \evaltarget{Satisfy all listed final states while preserving unrelated devices.}{device(5835).\allowbreak work\_\allowbreak state == playing and volume\_\allowbreak level == 40; device(1144).\allowbreak brightness == 30; device(5934).\allowbreak fan.\allowbreak state == on and fan.\allowbreak preset\_\allowbreak mode == natural} \\
\taxrowsep

\taxsub{Batch Operations} & ``Set all lights in the home to 30\% brightness and the warmest color temperature.'' & \evaltarget{Apply the same setting to every selected light.}{all selected lights: state == on, brightness == 30, color\_\allowbreak temperature == 2700} \\
\taxrowsep

\taxsub{State-dependent} & ``Set all lights whose brightness is above 80\% to 40\%.'' & \evaltarget{Only lights satisfying the brightness predicate are changed.}{device(2965).\allowbreak light.\allowbreak brightness == 40; device(6460).\allowbreak brightness == 40; device(4072).\allowbreak light.\allowbreak brightness == 40} \\
\taxrowsep

\taxsub{Room-dependent} & ``Except for rooms where activity is detected, turn off the lights in the other rooms and stop any speakers or smart screens still playing there.'' & \evaltarget{Preserve occupied rooms and update only unoccupied-room devices.}{device(6681).\allowbreak work\_\allowbreak state in \{standby, paused\}; device(7602).\allowbreak state == off; device(6022).\allowbreak state == off} \\
\taxrowsep

\taxsub{Top-N Selection} & ``Set the two fans with the highest wind speed to sleep mode and turn on oscillation for both.'' & \evaltarget{Select the two highest-speed fans and update both.}{device(5035).\allowbreak state == on and preset\_\allowbreak mode == sleep; device(5889).\allowbreak fan.\allowbreak state == on and fan.\allowbreak preset\_\allowbreak mode == sleep and fan.\allowbreak oscillation\_\allowbreak enabled == True} \\
\taxrowsep

\taxgroup{TC3: Ambiguous Intent}
\taxsub{State-level Ambiguity} & ``The room is too dark.'' & \evaltarget{Resolve or clarify the lighting intervention, then adjust the selected lights or curtains.}{device(3460).\allowbreak light.\allowbreak brightness == 80; device(2695).\allowbreak position == 100; device(8066).\allowbreak brightness == 80} \\
\taxrowsep

\taxsub{Scene-level Ambiguity} & ``I am going to rest for a while.'' & \evaltarget{Ground the scene in the appropriate smart-home action.}{device(9354).\allowbreak media\_\allowbreak name == light music and volume\_\allowbreak muted\_\allowbreak enabled == False} \\
\taxrowsep

\taxsub{Slot-missing} & ``Use the oven to heat up my lunch.'' & \evaltarget{Ask for the missing duration; after the user says five minutes, start reheating.}{device(4272).\allowbreak microwave\_\allowbreak steam\_\allowbreak oven.\allowbreak state == cooking; work\_\allowbreak mode == reheating; target\_\allowbreak cooking\_\allowbreak time == 300; remaining\_\allowbreak running\_\allowbreak time == 300} \\
\taxrowsep

\taxgroup{TC4: Automated Task Scheduling}
\taxsub{Time-specific} & ``At 10 tonight, lock the parents' bedroom door and enable defense mode; set the bathroom fan-light to the dimmest warm light and lowest fan speed.'' & \evalauto{Store the scheduled lock, light, and fan actions.}{time\_\allowbreak cron = 0 0 22 27 6 ? 2024}{device(8993).\allowbreak defend\_\allowbreak enabled == True and state == locked; device(9121).\allowbreak light.\allowbreak state == on and color\_\allowbreak temperature == 2700 and brightness == 1 and fan.\allowbreak state == on} \\
\taxrowsep

\taxsub{State-driven} & ``When the kitchen has had no movement for fifteen minutes, automatically turn off the foyer fan-light and set the living-room smart screen volume to 30.'' & \evalauto{Store the condition-action rule for the fan-light and smart screen.}{device(5267).\allowbreak no\_\allowbreak move\_\allowbreak duration > 900}{device(1870).\allowbreak light.\allowbreak state == off and fan.\allowbreak state == off; device(7813).\allowbreak volume\_\allowbreak level == 30} \\
\taxrowsep

\taxsub{Time-and-state} & ``Every day at 2 p.m., if the kitchen temperature is still above 35 degrees, start the living-room vacuum and set the bedroom fan-light fan to high.'' & \evalauto{Start the vacuum and set the fan to high only when both conditions hold.}{time\_\allowbreak cron = 0 0 14 * * ? *; device(8666).\allowbreak current\_\allowbreak temperature > 35.0}{device(2194).\allowbreak fan.\allowbreak speed\_\allowbreak level == high and device(3607).\allowbreak state == cleaning} \\
\taxrowsep

\taxgroup{TC5: Context-Aware Multi-turn Interaction}
\taxsub{Context Continuation} & Previous turn: rename the main-bedroom temperature-humidity sensor. Current turn: ``Change its alias to main-bedroom temperature-humidity sensor.'' & \evaltarget{Resolve ``its'' to the previous sensor.}{device(8431).\allowbreak alias == main-bedroom temperature-humidity sensor} \\
\taxrowsep

\taxsub{Topic Shifts} & Previous turn: play media on a smart screen. Current turn: ``Turn on the balcony fan-light in the master room, set the fan to natural mode, and set brightness to 50\%.'' & \evaltarget{Do not inherit the previous media target; execute the new fan-light request.}{device(3625).\allowbreak light.\allowbreak brightness == 50 and fan.\allowbreak preset\_\allowbreak mode == natural} \\
\taxrowsep

\taxsub{Coreference Resolution} & Previous turn: turn on the air cube in the entertainment room. Current turn: ``Turn on the speaker too.'' & \evaltarget{Resolve the speaker to the relevant room.}{device(8817).\allowbreak player.\allowbreak volume\_\allowbreak muted\_\allowbreak enabled == False and player.\allowbreak work\_\allowbreak state == playing} \\
\taxrowsep

\taxsub{User Corrections} & Previous turn: set the secondary-bedroom fan-light direction forward. Current turn: ``Do not adjust the secondary bedroom; turn on the boy's-room fan.'' & \evaltarget{Restore the previous fan-light setting and execute the corrected target.}{device(3024).\allowbreak fan.\allowbreak current\_\allowbreak direction == reverse and device(7481).\allowbreak state == on} \\
\taxrowsep

\taxgroup{TC6: Personalized Memory}
\taxsub{Short-term Memory} & ``Do not adjust the living-room air cube after all; turn on the guest-bathroom fan instead.'' & \evaltarget{Undo the recent air-cube adjustment and apply the new fan action.}{device(7322).\allowbreak state == off and target\_\allowbreak humidity == 46 and target\_\allowbreak temperature == 11; device(6899).\allowbreak fan.\allowbreak state == on} \\
\taxrowsep

\taxsub{Long-term Memory} & ``Time to sleep.'' & \taxblock{Target}{Retrieve the user's sleep preference and apply it.}\par\vspace{5pt}\taxblock{memory\_list}{The user prefers the air cube to set target humidity to 50\%, turn off the breathing light, and enter fall-asleep mode when they say it is time to sleep.}\par\vspace{5pt}\taxcodeblock{Critic}{device(3415).\allowbreak sleep\_\allowbreak state == fall\_\allowbreak asleep and breathing\_\allowbreak light\_\allowbreak enabled == False and state == on and target\_\allowbreak humidity == 50} \\
\taxrowsep

\taxgroup{TC7: Environment-Grounded Query}
\taxsub{Status Inquiries} & ``What are the current states of the vacuums in the secondary bedroom and guest bathroom?'' & \evalquery{Answer without changing any device state.}{Both vacuums are in standby and are not currently cleaning.} \\
\taxrowsep

\taxsub{Quantity Queries} & ``How many fan-lights currently have both the light and fan turned on?'' & \evalquery{Count the devices satisfying both predicates without changing state.}{Zero fan-lights satisfy both predicates in the sampled record.} \\

\end{longtable}
\endgroup

%% file: appendix/implementation_details_for_the_two_evaluation_settings_backup7.tex
\section{Implementation Details for the Two Evaluation Settings}
\label{app:evaluation_settings}

This appendix gives the exact implementation of the two evaluation formats used in SMH-Bench. Both formats start from the same HomeEnv state and are scored by the same verifier. The only difference is the interface given to the model: \emph{Direct Reasoning} is a full-context, one-shot prediction format, while \emph{Environment-Interactive Agent} is a partial-observation, tool-mediated interaction format.

\subsection{Two Evaluation Formats}
\label{app:direct_setup}
\label{app:tool_setup}

\begingroup
\small
\setlength{\tabcolsep}{5pt}
\renewcommand{\arraystretch}{1.18}
\newcommand{\formatcell}[2]{\textbf{#1}\par\smallskip #2}
\newcommand{\formatcode}[1]{{\ttfamily\footnotesize #1}}
\begin{longtable}{>{\raggedright\arraybackslash}p{0.18\linewidth}
                >{\raggedright\arraybackslash}p{0.37\linewidth}
                >{\raggedright\arraybackslash}p{0.37\linewidth}}
\caption{Two SMH-Bench evaluation formats.}
\label{tab:two_evaluation_formats}\\
\toprule
Interface component & Direct Reasoning & Environment-Interactive Agent \\
\midrule
\endfirsthead
\toprule
Interface component & Direct Reasoning & Environment-Interactive Agent \\
\midrule
\endhead
\multicolumn{3}{r}{Continued on next page} \\
\midrule
\endfoot
\bottomrule
\endlastfoot

\multicolumn{3}{l}{\textbf{Model-facing interface}} \\
\midrule
\textbf{Initial observability} &
\formatcell{Full context.}{Request, dialogue history, memory, entrance device, time, rooms, devices, current values, component paths, service schemas, arguments, ranges, enums, and deduplicated device-type schemas} &
\formatcell{Partial context.}{Request, dialogue history, memory, entrance device, time, room list, and device index. Exact states, schemas, and locators must be obtained through tools.} \\
\midrule
\textbf{Runtime interface} &
\formatcell{One-shot.}{The model produces one final answer for the stage. It cannot query HomeEnv, observe execution feedback, or repair an invalid locator before replay.} &
\formatcell{Tool loop.}{The model interacts with HomeEnv through \formatcode{query\_device} for candidates, specs, and status, and \formatcode{control\_device} for immediate service execution. Observations are appended before the next step.} \\
\midrule
\textbf{Model output} &
\formatcell{Structured final JSON.}{Execution uses \formatcode{\{mode, response, actions\}}, where each action has \formatcode{\{did, locator, arguments\}}. Clarification uses \formatcode{\{mode, response\}}. Query-only tasks use \formatcode{response}.} &
\formatcell{Tool calls plus final text.}{Intermediate steps are OpenAI-style calls to \formatcode{query\_device} with \formatcode{\{what, did, room, tags, ...\}} or \formatcode{control\_device} with \formatcode{\{did, locator, arguments\}}. The final answer is natural language after tool use stops.} \\
\midrule
\multicolumn{3}{l}{\textbf{Evaluator-side extraction and scoring}} \\
\midrule
\textbf{Prediction extraction} &
\formatcell{Replay.}{Parse the JSON, replay \formatcode{actions} in a fresh HomeEnv copy, and convert the replayed state delta into \formatcode{predictions.labels}.} &
\formatcell{State difference}{Snapshot HomeEnv before and after the tool loop. The environment delta becomes \formatcode{predictions.labels}; the final assistant text becomes \formatcode{predictions.response}.} \\
\midrule
\textbf{Main capability tested} &
One-shot grounding from complete serialized context: target selection, component resolution, action formatting, argument typing, and no-op avoidance. &
Interactive deployment behavior under partial observability: evidence gathering, schema discovery, tool sequencing, execution feedback handling, and stopping after the environment satisfies the request. \\
\midrule
\textbf{Same final verifier} &
\multicolumn{2}{>{\raggedright\arraybackslash}p{0.74\linewidth}}{After either format, SMH-Bench applies the same record-level checks: critic evaluation for state-changing tasks, query-response judging for query tasks, clarification alignment for clarification tasks, token accounting, and final \formatcode{overall\_pass}.} \\

\end{longtable}
\endgroup

\FloatBarrier

\subsection{Evaluation Pipeline}

SMH-Bench uses a unified evaluation pipeline for both settings. Given a set of benchmark samples, the pipeline either loads existing predictions or runs model inference under one of the two interfaces. In the Direct Reasoning setting, the model receives the serialized home context and produces a structured final prediction, which is replayed in a fresh HomeEnv instance to obtain the predicted state changes. In the Environment-Interactive Agent setting, the model observes a partial home context, interacts with HomeEnv through tools, and the resulting environment delta is used as the prediction.

After inference, both settings are normalized into the same prediction schema, including predicted state labels, optional natural-language responses, automation conditions, dialogue traces, token usage, and execution errors. The shared verifier then evaluates each sample according to its task type: state-changing tasks are checked by the critic over the predicted home state, query-only tasks are judged by response quality, clarification tasks require the appropriate clarification behavior, and automation tasks additionally check predicted trigger or condition semantics. A sample passes only when all required checks for that sample type pass. Finally, SMH-Bench reports record-level evaluation results and aggregate pass rates by task category and evaluation setting.

\FloatBarrier

\subsection{Direct Reasoning Trajectory and Environment-Interactive Agent Trajectory}

\begin{promptformatbox}{Direct Reasoning Trajectory}
Query:
  Turn on all kitchen-area lights that are currently off, and set their brightness to 70

Observation:
  The model receives full home context before generation.
  Relevant excerpt:
    room 11 = kitchen area
    did 9313 = fan-light, tags=[fan, light], room=11
      state: light.state=off, light.brightness=32
      services: light.turn_on(), light.set_brightness(brightness:int[1,100])
    did 6152 = light, tags=[light], room=11
      state: state=off, brightness=33
      services: turn_on(), set_brightness(brightness:int[1,100])
    did 2284 = light, tags=[light], room=11
      state: state=off, brightness=16
      services: turn_on(), set_brightness(brightness:int[1,100])
    did 8381 = light, tags=[light], room=11
      state: state=off, brightness=16
      services: turn_on(), set_brightness(brightness:int[1,100])

Example Trajectory:
  Step 1. The model directly emits the final structured execution payload:
  {
    "mode": "execution",
    "response": "The kitchen-area lights that were off are now on and set to 70
    "actions": [
      {"did": "9313", "locator": "light.turn_on", "arguments": {}},
      {"did": "9313", "locator": "light.set_brightness", "arguments": {"brightness": 70}},
      {"did": "6152", "locator": "turn_on", "arguments": {}},
      {"did": "6152", "locator": "set_brightness", "arguments": {"brightness": 70}},
      {"did": "2284", "locator": "turn_on", "arguments": {}},
      {"did": "2284", "locator": "set_brightness", "arguments": {"brightness": 70}},
      {"did": "8381", "locator": "turn_on", "arguments": {}},
      {"did": "8381", "locator": "set_brightness", "arguments": {"brightness": 70}}
    ]
  }

Replay:
  The evaluator starts from a fresh copy of engine_0 and executes the predicted actions.
    action 1: 9313.light.turn_on()           -> 9313.light.state: off -> on
    action 2: 9313.light.set_brightness(70)  -> 9313.light.brightness: 32 -> 70
    action 3: 6152.turn_on()                 -> 6152.state: off -> on
    action 4: 6152.set_brightness(70)        -> 6152.brightness: 33 -> 70
    action 5: 2284.turn_on()                 -> 2284.state: off -> on
    action 6: 2284.set_brightness(70)        -> 2284.brightness: 16 -> 70
    action 7: 8381.turn_on()                 -> 8381.state: off -> on
    action 8: 8381.set_brightness(70)        -> 8381.brightness: 16 -> 70

Prediction:
  labels = [
    {"did":"9313","attribute":"light.state","value":"on"},
    {"did":"9313","attribute":"light.brightness","value":70},
    {"did":"6152","attribute":"state","value":"on"},
    {"did":"6152","attribute":"brightness","value":70},
    {"did":"2284","attribute":"state","value":"on"},
    {"did":"2284","attribute":"brightness","value":70},
    {"did":"8381","attribute":"state","value":"on"},
    {"did":"8381","attribute":"brightness","value":70}
  ]

Verification:
  Critic =
    device('9313').light.state == "on" and device('9313').light.brightness == 70 and
    device('6152').state == "on" and device('6152').brightness == 70 and
    device('2284').state == "on" and device('2284').brightness == 70 and
    device('8381').state == "on" and device('8381').brightness == 70
  Result: TRUE. The sample passes under Direct Reasoning.
\end{promptformatbox}

\FloatBarrier

\begin{promptformatbox}{Environment-Interactive Agent Trajectory}
Query:
  Turn on all kitchen-area lights that are currently off, and set their brightness to 70

Observation:
  The model receives only brief home context before acting.
  Relevant excerpt:
    room 11 = kitchen area
    device index in room 11:
      did 9313 = fan-light, tags=[fan, light]
      did 6152 = light, tags=[light]
      did 2284 = light, tags=[light]
      did 8381 = light, tags=[light]
  Exact attributes, service locators, and argument schemas are not assumed; they are queried.

Example Trajectory:
  Step 1. query_device({"what":"brief","room":"kitchen area","tags":"light"})
    Observation -> candidates: 9313, 6152, 2284, 8381.

  Step 2. query_device({"what":"spec_status","did":"9313"})
    Observation -> fan-light; light.state=off; light.brightness=32;
                   valid light services: light.turn_on(),
                   light.set_brightness(brightness:int[1,100]).
    Meaning     -> component namespace is required; plain turn_on would not target the light.

  Step 3. query_device({"what":"spec_status","did":"6152"})
    Observation -> device-level light; state=off; brightness=33;
                   valid services: turn_on(), set_brightness(brightness:int[1,100]).

  Step 4. query_device({"what":"spec_status","did":"2284"})
    Observation -> device-level light; state=off; brightness=16;
                   valid services: turn_on(), set_brightness(brightness:int[1,100]).

  Step 5. query_device({"what":"spec_status","did":"8381"})
    Observation -> device-level light; state=off; brightness=16;
                   valid services: turn_on(), set_brightness(brightness:int[1,100]).

  Step 6. control_device({"did":"9313","locator":"light.turn_on","arguments":{}})
    Observation -> ok=true; 9313.light.state: off -> on.

  Step 7. control_device({"did":"9313","locator":"light.set_brightness","arguments":{"brightness":70}})
    Observation -> ok=true; 9313.light.brightness: 32 -> 70.

  Step 8. control_device({"did":"6152","locator":"turn_on","arguments":{}})
    Observation -> ok=true; 6152.state: off -> on.

  Step 9. control_device({"did":"6152","locator":"set_brightness","arguments":{"brightness":70}})
    Observation -> ok=true; 6152.brightness: 33 -> 70.

  Step 10. control_device({"did":"2284","locator":"turn_on","arguments":{}})
    Observation -> ok=true; 2284.state: off -> on.

  Step 11. control_device({"did":"2284","locator":"set_brightness","arguments":{"brightness":70}})
    Observation -> ok=true; 2284.brightness: 16 -> 70.

  Step 12. control_device({"did":"8381","locator":"turn_on","arguments":{}})
    Observation -> ok=true; 8381.state: off -> on.

  Step 13. control_device({"did":"8381","locator":"set_brightness","arguments":{"brightness":70}})
    Observation -> ok=true; 8381.brightness: 16 -> 70.

  Step 14. Final response:
    "Done. I turned on the kitchen-area lights that were off and set them to 70

Prediction:
  The evaluator does not parse labels from the final text.
  It computes labels by diffing the HomeEnv snapshot before Step 1 and after Step 14:
    {"did":"9313","attribute":"light.state","value":"on"}
    {"did":"9313","attribute":"light.brightness","value":70}
    {"did":"6152","attribute":"state","value":"on"}
    {"did":"6152","attribute":"brightness","value":70}
    {"did":"2284","attribute":"state","value":"on"}
    {"did":"2284","attribute":"brightness","value":70}
    {"did":"8381","attribute":"state","value":"on"}
    {"did":"8381","attribute":"brightness","value":70}

Verification:
  The same critic used above evaluates to TRUE.
  Result: TRUE. The sample passes under Environment-Interactive Agent.
\end{promptformatbox}

\FloatBarrier

%% file: appendix/error_distribution_breakdown.tex
\section{Error Distribution Breakdown}
\label{app:error-distribution}

\subsection{Complete Error Distribution}

Table~\ref{tab:error_distribution_full} reports the full error breakdown for the four representative models analyzed in the main experiments. Counts are grouped by evaluation format: Environment-Interactive Agent (EIA) and Direct Reasoning (DR). Each cell reports the number of categorized errors, with the percentage share within that model and setting in parentheses. A dash indicates that the error type was not observed in the categorized failures for that model.

\begingroup
\scriptsize
\setlength{\tabcolsep}{3.5pt}
\renewcommand{\arraystretch}{1.12}
\begin{longtable}{>{\raggedright\arraybackslash}p{0.12\linewidth}
                  >{\raggedright\arraybackslash}p{0.31\linewidth}
                  >{\centering\arraybackslash}p{0.115\linewidth}
                  >{\centering\arraybackslash}p{0.115\linewidth}
                  >{\centering\arraybackslash}p{0.115\linewidth}
                  >{\centering\arraybackslash}p{0.115\linewidth}}
\caption{Complete breakdown of categorized error types and per-model frequencies. Percentages are computed within each model-setting error distribution.}
\label{tab:error_distribution_full}\\
\toprule
Setting & Error type & Gemini-3.1-Pro & Claude-Sonnet-4.6 & GPT-5.4 & DeepSeek-V3.2 \\
\midrule
\endfirsthead
\toprule
Setting & Error type & Gemini-3.1-Pro & Claude-Sonnet-4.6 & GPT-5.4 & DeepSeek-V3.2 \\
\midrule
\endhead
\midrule
\multicolumn{6}{r}{Continued on next page} \\
\midrule
\endfoot
\bottomrule
\endlastfoot

\multicolumn{6}{l}{\textbf{Environment-Interactive Agent (EIA)}} \\
\midrule
Structured output & Missing labels & 44 (31.2\%) & 53 (37.6\%) & 116 (58.3\%) & 113 (43.3\%) \\
Structured output & Missing automation conditions & 13 (9.2\%) & 6 (4.3\%) & 44 (22.1\%) & 17 (6.5\%) \\
Structured output & Missing query response & -- & -- & -- & 2 (0.8\%) \\
Structured output & Empty model output & -- & -- & 5 (2.5\%) & -- \\
\midrule
Automation judging & Conditions judge error & 18 (12.8\%) & 15 (10.6\%) & 18 (9.0\%) & 14 (5.4\%) \\
\midrule
Tool-call validity & Missing required tool arguments & 33 (23.4\%) & 56 (39.7\%) & 9 (4.5\%) & 16 (6.1\%) \\
Tool-call validity & Invalid \texttt{None} value in tool arguments & 32 (22.7\%) & 8 (5.7\%) & -- & 6 (2.3\%) \\
Tool-call validity & Integer passed where string was required & -- & -- & -- & 11 (4.2\%) \\
Tool-call validity & Other tool or schema errors & 3 (2.1\%) & 3 (2.1\%) & 4 (2.0\%) & 5 (1.9\%) \\
\midrule
Interaction efficiency & Too many tool calls & -- & -- & -- & 75 (28.7\%) \\

\midrule
\multicolumn{6}{l}{\textbf{Direct Reasoning (DR)}} \\
\midrule
Structured output & Missing labels & 50 (37.3\%) & 75 (29.3\%) & 96 (44.7\%) & 135 (33.6\%) \\
Structured output & Missing automation conditions & 12 (9.0\%) & 10 (3.9\%) & 26 (12.1\%) & 20 (5.0\%) \\
Structured output & Missing query response & -- & 8 (3.1\%) & -- & -- \\
\midrule
Mode selection & Execution predicted as clarification & 33 (24.6\%) & 14 (5.5\%) & 60 (27.9\%) & 85 (21.1\%) \\
Mode selection & Clarification predicted as execution & 32 (23.9\%) & 46 (18.0\%) & 25 (11.6\%) & 26 (6.5\%) \\
\midrule
Automation judging & Conditions judge error & 4 (3.0\%) & 4 (1.6\%) & 4 (1.9\%) & 7 (1.7\%) \\
\midrule
Action/schema validity & Missing required action argument & 2 (1.5\%) & 2 (0.8\%) & 2 (0.9\%) & -- \\
Action/schema validity & Invalid \texttt{None} value in action arguments & -- & 33 (12.9\%) & -- & -- \\
Action/schema validity & Wrong action argument name & -- & -- & -- & 73 (18.2\%) \\
Action/schema validity & Other action execution error & -- & -- & -- & 24 (6.0\%) \\
\midrule
Format/runtime & LLM request or JSON parsing error & 1 (0.7\%) & 58 (22.7\%) & 2 (0.9\%) & 24 (6.0\%) \\

\end{longtable}
\endgroup

\subsection{Representative Error Case Studies}
\label{app:error-case-study}

Table~\ref{tab:representative_error_cases} summarizes three representative failures that correspond to the dominant categories in Table~\ref{tab:error_distribution_full}. We report each case by its task requirement, observed behavior, evaluator-visible failure, and underlying mechanism. This format preserves the diagnostic evidence from the recorded runs while omitting identifiers and low-level trace fields that do not affect the interpretation.

\begin{table}[t]
\centering
\small
\setlength{\tabcolsep}{3.5pt}
\renewcommand{\arraystretch}{1.12}
\caption{Representative failure mechanisms observed in the categorized errors.}
\label{tab:representative_error_cases}
\begin{tabular}{>{\raggedright\arraybackslash}p{0.17\linewidth}
                >{\raggedright\arraybackslash}p{0.25\linewidth}
                >{\raggedright\arraybackslash}p{0.28\linewidth}
                >{\raggedright\arraybackslash}p{0.22\linewidth}}
\toprule
Case & Task requirement & Observed behavior & Diagnosis \\
\midrule
Schema grounding & Start a water heater using a service with required arguments. & The model selected the correct device and service but called \texttt{start\_heating} with only \texttt{work\_mode}. & Correct semantic grounding, incomplete service-signature instantiation. \\
Automation serialization & Create a presence-triggered automation that turns on stair and corridor lights at half brightness. & The model inspected candidate sensors but produced no \texttt{create\_automation} call, no conditions, and no action labels. & High-level automation intent recognized, but not serialized into executable artifacts. \\
Interaction control & Configure a reception-room hosting scene under a 10-call interaction budget. & The model over-queried devices and exhausted the budget after only partial lighting changes. & Inefficient tool policy left the multi-device request incomplete. \\
\bottomrule
\end{tabular}
\end{table}

\paragraph{Case 1: Schema-grounding failure in single-device control.}
The user asked the agent to start heating the electric water heater in the secondary-bedroom public bathroom. The agent resolved the correct device, device 3981, and inspected the service schema for \texttt{start\_heating}, which required \texttt{work\_mode}, \texttt{target\_water\_quantity}, and \texttt{target\_water\_temperature}. However, the subsequent control call supplied only \texttt{work\_mode=quick}. The environment rejected the call with a missing-argument error, and the final prediction contained no state-change labels. This case shows a schema-grounding failure: the model identified the right device and operation but failed to instantiate the complete executable signature.

\paragraph{Case 2: Automation serialization failure.}
The user requested an automation that turns on stair and corridor lights at half brightness whenever human-presence sensors in the specified areas detect a person. The model recognized the relevant trigger family and repeatedly inspected candidate sensors, but it stopped by asking for confirmation about partial sensor availability. It never issued a \texttt{create\_automation} call and emitted neither \texttt{conditions} nor \texttt{labels}, which made both the trigger and effect unevaluable. This case illustrates that automation errors can occur even when the high-level intent is understood: the missing step is the serialization of the rule into concrete condition and action artifacts.

\paragraph{Case 3: Interaction-control failure under a tool budget.}
In an ambiguous hosting-scene task, the user clarified that the reception room should feel warm and welcoming by adjusting lights, air conditioning, and music. After finding four reception-room lights, the model spent six additional calls inspecting individual devices before executing any action. It then completed only one light's state, brightness, and color-temperature changes before exceeding the 10-call budget, leaving the climate and music subgoals unfinished. This case is primarily an interaction-control failure rather than a semantic failure: the model understood the scene goal, but its tool-use policy consumed the budget before the multi-device task was complete.

%% file: appendix/detailed_model_results.tex
\section{Detailed Model Performance Results}
\label{app:detailed-results}

This appendix reports the complete success-rate breakdowns underlying the main experimental table. DR denotes the Direct Reasoning setting, and EIA denotes the Environment-Interactive Agent setting. TC1--TC7 follow the taxonomy in Section~\ref{sec:TC}.

\subsection{Full Pass Rates by Difficulty}
\label{app:full-pass-rates}

Table~\ref{tab:full_pass_rate_by_difficulty} expands Table~\ref{tab:main_results} by retaining the home-complexity split. Each capability cell reports success rate in percent.

\begingroup
\scriptsize
\setlength{\tabcolsep}{1.5pt}
\renewcommand{\arraystretch}{1.08}
\begin{longtable}{>{\raggedright\arraybackslash}p{0.145\linewidth}>{\centering\arraybackslash}p{0.052\linewidth}>{\centering\arraybackslash}p{0.070\linewidth}>{\centering\arraybackslash}p{0.055\linewidth}>{\centering\arraybackslash}p{0.079\linewidth}>{\centering\arraybackslash}p{0.079\linewidth}>{\centering\arraybackslash}p{0.079\linewidth}>{\centering\arraybackslash}p{0.079\linewidth}>{\centering\arraybackslash}p{0.079\linewidth}>{\centering\arraybackslash}p{0.079\linewidth}>{\centering\arraybackslash}p{0.079\linewidth}}
\caption{Complete final pass rates by model, evaluation setting, home difficulty, and task category. }
\label{tab:full_pass_rate_by_difficulty}\\
\toprule
Model & Setting & Difficulty & Avg & \shortstack{TC1\\Atom} & \shortstack{TC2\\Comp.} & \shortstack{TC3\\Ambig.} & \shortstack{TC4\\Auto.} & \shortstack{TC5\\Multi.} & \shortstack{TC6\\Memory} & \shortstack{TC7\\Query} \\
\midrule
\endfirsthead
\toprule
Model & Setting & Difficulty & Avg & \shortstack{TC1\\Atom} & \shortstack{TC2\\Comp.} & \shortstack{TC3\\Ambig.} & \shortstack{TC4\\Auto.} & \shortstack{TC5\\Multi.} & \shortstack{TC6\\Memory} & \shortstack{TC7\\Query} \\
\midrule
\endhead
\midrule
\multicolumn{11}{r}{Continued on next page} \\
\midrule
\endfoot
\bottomrule
\endlastfoot

Qwen3.5-4B & DR & Simple & 55.1 & 73.3 & 48.0 & 60.0 & 34.7 & 53.0 & 46.0 & 82.0 \\
 & DR & Medium & 43.6 & 73.3 & 30.7 & 33.3 & 17.8 & 51.7 & 33.3 & 80.0 \\
 & DR & Complex & 26.4 & 50.0 & 12.0 & 23.3 & 6.7 & 30.0 & 30.0 & 50.0 \\

 & EIA & Simple & 65.8 & 77.3 & 60.0 & 85.3 & 25.3 & 72.0 & 64.0 & 84.0 \\
 & EIA & Medium & 48.8 & 68.9 & 37.3 & 64.4 & 20.0 & 51.7 & 43.3 & 66.7 \\
 & EIA & Complex & 44.5 & 50.0 & 22.0 & 50.0 & 13.3 & 67.5 & 60.0 & 70.0 \\
\midrule
Qwen3.5-9B & DR & Simple & 59.8 & 82.7 & 52.8 & 60.0 & 34.7 & 56.0 & 60.0 & 88.0 \\
 & DR & Medium & 41.8 & 66.7 & 29.3 & 31.1 & 20.0 & 50.0 & 33.3 & 76.7 \\
 & DR & Complex & 33.6 & 60.0 & 14.0 & 40.0 & 10.0 & 42.5 & 35.0 & 50.0 \\

 & EIA & Simple & 70.0 & 78.7 & 67.2 & 84.0 & 46.7 & 74.0 & 60.0 & 80.0 \\
 & EIA & Medium & 62.1 & 71.1 & 52.0 & 64.4 & 24.4 & 80.0 & 63.3 & 90.0 \\
 & EIA & Complex & 53.2 & 80.0 & 34.0 & 40.0 & 26.7 & 65.0 & 80.0 & 70.0 \\
\midrule
Qwen3.5-397B & DR & Simple & 84.2 & 94.7 & 88.0 & 85.3 & 66.7 & 86.0 & 72.0 & 92.0 \\
 & DR & Medium & 73.0 & 93.3 & 64.0 & 62.2 & 57.8 & 83.3 & 76.7 & 80.0 \\
 & DR & Complex & 67.7 & 86.7 & 44.0 & 76.7 & 46.7 & 85.0 & 80.0 & 70.0 \\

 & EIA & Simple & 84.9 & 93.3 & 83.2 & 90.7 & 72.0 & 87.0 & 74.0 & 94.0 \\
 & EIA & Medium & 79.4 & 86.7 & 62.7 & 82.2 & 75.6 & 83.3 & 83.3 & 100.0 \\
 & EIA & Complex & 57.7 & 76.7 & 42.0 & 83.3 & 53.3 & 30.0 & 75.0 & 75.0 \\
\midrule
Qwen3.5-Plus & DR & Simple & 75.1 & 92.0 & 71.2 & 72.0 & 64.0 & 79.0 & 66.0 & 82.0 \\
 & DR & Medium & 53.3 & 64.4 & 44.0 & 28.9 & 42.2 & 65.0 & 76.7 & 66.7 \\
 & DR & Complex & 45.0 & 70.0 & 14.0 & 33.3 & 26.7 & 75.0 & 75.0 & 40.0 \\

 & EIA & Simple & 76.9 & 92.0 & 75.2 & 85.3 & 53.3 & 80.0 & 60.0 & 92.0 \\
 & EIA & Medium & 67.6 & 95.6 & 49.3 & 71.1 & 33.3 & 78.3 & 80.0 & 83.3 \\
 & EIA & Complex & 56.4 & 83.3 & 24.0 & 63.3 & 30.0 & 80.0 & 55.0 & 80.0 \\
\midrule
DeepSeek-V3.2 & DR & Simple & 79.8 & 85.3 & 81.6 & 78.7 & 61.3 & 85.0 & 70.0 & 96.0 \\
 & DR & Medium & 62.7 & 71.1 & 45.3 & 55.6 & 53.3 & 76.7 & 63.3 & 90.0 \\
 & DR & Complex & 53.6 & 76.7 & 32.0 & 46.7 & 26.7 & 72.5 & 70.0 & 70.0 \\

 & EIA & Simple & 85.6 & 90.7 & 80.8 & 98.7 & 74.7 & 88.0 & 74.0 & 94.0 \\
 & EIA & Medium & 68.8 & 84.4 & 32.0 & 77.8 & 62.2 & 83.3 & 73.3 & 100.0 \\
 & EIA & Complex & 57.7 & 83.3 & 14.0 & 63.3 & 40.0 & 80.0 & 75.0 & 85.0 \\
\midrule
MiniMax-M2.7 & DR & Simple & 64.0 & 88.0 & 76.8 & 74.7 & 52.0 & 39.0 & 32.0 & 80.0 \\
 & DR & Medium & 45.8 & 77.8 & 44.0 & 51.1 & 28.9 & 28.3 & 40.0 & 60.0 \\
 & DR & Complex & 30.5 & 70.0 & 20.0 & 40.0 & 20.0 & 22.5 & 20.0 & 25.0 \\

 & EIA & Simple & 45.8 & 64.0 & 17.6 & 65.3 & 52.0 & 34.0 & 22.0 & 98.0 \\
 & EIA & Medium & 45.2 & 42.2 & 37.3 & 40.0 & 42.2 & 48.3 & 26.7 & 93.3 \\
 & EIA & Complex & 38.2 & 63.3 & 18.0 & 36.7 & 16.7 & 47.5 & 40.0 & 65.0 \\
\midrule
GLM-5 & DR & Simple & 72.9 & 90.7 & 72.0 & 77.3 & 68.0 & 71.0 & 34.0 & 92.0 \\
 & DR & Medium & 50.9 & 75.6 & 38.7 & 20.0 & 40.0 & 68.3 & 46.7 & 76.7 \\
 & DR & Complex & 40.0 & 63.3 & 12.0 & 23.3 & 30.0 & 75.0 & 40.0 & 45.0 \\

 & EIA & Simple & 76.9 & 94.7 & 78.4 & 92.0 & 52.0 & 76.0 & 46.0 & 94.0 \\
 & EIA & Medium & 67.3 & 84.4 & 62.7 & 82.2 & 31.1 & 70.0 & 60.0 & 86.7 \\
 & EIA & Complex & 59.5 & 80.0 & 38.0 & 63.3 & 36.7 & 85.0 & 50.0 & 70.0 \\
\midrule
GPT-5.4-Mini & DR & Simple & 77.3 & 89.3 & 81.6 & 77.3 & 53.3 & 81.0 & 58.0 & 96.0 \\
 & DR & Medium & 63.9 & 71.1 & 57.3 & 48.9 & 42.2 & 81.7 & 66.7 & 86.7 \\
 & DR & Complex & 54.1 & 73.3 & 40.0 & 40.0 & 30.0 & 70.0 & 65.0 & 75.0 \\

 & EIA & Simple & 81.8 & 90.7 & 83.2 & 93.3 & 60.0 & 83.0 & 68.0 & 92.0 \\
 & EIA & Medium & 74.5 & 91.1 & 60.0 & 84.4 & 55.6 & 80.0 & 70.0 & 93.3 \\
 & EIA & Complex & 60.0 & 83.3 & 46.0 & 70.0 & 20.0 & 77.5 & 70.0 & 60.0 \\
\midrule
GPT-5.4 & DR & Simple & 85.6 & 92.0 & 90.4 & 84.0 & 73.3 & 89.0 & 66.0 & 98.0 \\
 & DR & Medium & 77.3 & 93.3 & 68.0 & 64.4 & 66.7 & 83.3 & 80.0 & 96.7 \\
 & DR & Complex & 74.5 & 86.7 & 66.0 & 83.3 & 43.3 & 80.0 & 90.0 & 85.0 \\

 & EIA & Simple & 85.1 & 94.7 & 88.8 & 92.0 & 56.0 & 91.0 & 70.0 & 98.0 \\
 & EIA & Medium & 76.1 & 93.3 & 69.3 & 77.8 & 46.7 & 83.3 & 80.0 & 90.0 \\
 & EIA & Complex & 70.9 & 93.3 & 64.0 & 76.7 & 36.7 & 80.0 & 75.0 & 75.0 \\
\midrule
Gemini-3.1-Flash & DR & Simple & 78.5 & 90.7 & 86.4 & 66.7 & 70.7 & 74.0 & 62.0 & 96.0 \\
 & DR & Medium & 70.6 & 88.9 & 61.3 & 57.8 & 62.2 & 70.0 & 76.7 & 93.3 \\
 & DR & Complex & 67.7 & 86.7 & 60.0 & 63.3 & 43.3 & 72.5 & 85.0 & 75.0 \\

 & EIA & Simple & 81.8 & 89.3 & 85.6 & 90.7 & 56.0 & 82.0 & 70.0 & 98.0 \\
 & EIA & Medium & 74.2 & 84.4 & 62.7 & 84.4 & 48.9 & 81.7 & 73.3 & 96.7 \\
 & EIA & Complex & 62.3 & 83.3 & 40.0 & 70.0 & 30.0 & 72.5 & 80.0 & 85.0 \\
\midrule
Gemini-3.1-Pro & DR & Simple & 88.9 & 96.0 & 94.4 & 85.3 & 77.3 & 92.0 & 76.0 & 94.0 \\
 & DR & Medium & 78.8 & 95.6 & 69.3 & 53.3 & 68.9 & 83.3 & 100.0 & 100.0 \\
 & DR & Complex & 76.8 & 93.3 & 66.0 & 70.0 & 56.7 & 82.5 & 100.0 & 85.0 \\

 & EIA & Simple & 90.2 & 96.0 & 95.2 & 98.7 & 69.3 & 93.0 & 78.0 & 94.0 \\
 & EIA & Medium & 81.2 & 91.1 & 73.3 & 86.7 & 60.0 & 83.3 & 96.7 & 90.0 \\
 & EIA & Complex & 78.6 & 93.3 & 74.0 & 83.3 & 56.7 & 85.0 & 90.0 & 70.0 \\
\midrule
Claude-Haiku-4.5 & DR & Simple & 84.0 & 93.3 & 88.0 & 81.3 & 76.0 & 85.0 & 76.0 & 82.0 \\
 & DR & Medium & 71.5 & 86.7 & 61.3 & 64.4 & 60.0 & 76.7 & 80.0 & 83.3 \\
 & DR & Complex & 65.9 & 86.7 & 54.0 & 63.3 & 40.0 & 82.5 & 75.0 & 65.0 \\

 & EIA & Simple & 81.8 & 89.3 & 87.2 & 92.0 & 54.7 & 82.0 & 68.0 & 96.0 \\
 & EIA & Medium & 73.9 & 86.7 & 65.3 & 80.0 & 44.4 & 81.7 & 80.0 & 90.0 \\
 & EIA & Complex & 65.5 & 86.7 & 52.0 & 76.7 & 36.7 & 77.5 & 75.0 & 60.0 \\
\midrule
Claude-Sonnet-4.6 & DR & Simple & 82.2 & 90.7 & 86.4 & 76.0 & 73.3 & 84.0 & 70.0 & 90.0 \\
 & DR & Medium & 73.0 & 91.1 & 61.3 & 44.4 & 71.1 & 85.0 & 86.7 & 83.3 \\
 & DR & Complex & 68.6 & 90.0 & 58.0 & 50.0 & 53.3 & 82.5 & 80.0 & 75.0 \\

 & EIA & Simple & 88.0 & 94.7 & 93.6 & 93.3 & 72.0 & 86.0 & 74.0 & 98.0 \\
 & EIA & Medium & 81.2 & 93.3 & 70.7 & 86.7 & 68.9 & 83.3 & 86.7 & 90.0 \\
 & EIA & Complex & 78.6 & 90.0 & 66.0 & 93.3 & 60.0 & 85.0 & 85.0 & 80.0 \\

\end{longtable}
\endgroup
\subsection{DeepSeek-V3.2 Thinking Ablation}
\label{app:deepseek-thinking-ablation}

Table~\ref{tab:deepseek_thinking_vs_nonthinking_full} reports the complete DeepSeek-V3.2 ablation between the thinking run and the non-thinking run.Diff is computed as Non-Thinking minus Thinking, so negative values indicate degradation after removing thinking.

\begingroup
\scriptsize
\setlength{\tabcolsep}{4pt}
\renewcommand{\arraystretch}{1.08}
\begin{longtable}{>{\centering\arraybackslash}p{0.09\linewidth}>{\raggedright\arraybackslash}p{0.36\linewidth}>{\centering\arraybackslash}p{0.11\linewidth}>{\centering\arraybackslash}p{0.13\linewidth}>{\centering\arraybackslash}p{0.10\linewidth}}
\caption{DeepSeek-V3.2 thinking versus non-thinking success rates by evaluation setting, difficulty, and task category.}
\label{tab:deepseek_thinking_vs_nonthinking_full}\\
\toprule
Difficulty & Category & Thinking & Non-thinking & Diff \\
\midrule
\endfirsthead
\toprule
Difficulty & Category & Thinking & Non-thinking & Diff \\
\midrule
\endhead
\midrule
\multicolumn{5}{r}{Continued on next page} \\
\midrule
\endfoot
\bottomrule
\endlastfoot

\multicolumn{5}{l}{\textbf{EIA}} \\
\midrule
Simple & Ambiguous Intent & 0.9867 & 0.8667 & -0.1200 \\
Simple & Atomic Control & 0.9067 & 0.7333 & -0.1733 \\
Simple & Automated Task Scheduling & 0.7467 & 0.6667 & -0.0800 \\
Simple & Compositional Control & 0.8080 & 0.7280 & -0.0800 \\
Simple & Personalized Memory & 0.7400 & 0.4800 & \textbf{-0.2600} \\
Simple & Context-Aware Multi-turn Interaction & 0.8800 & 0.7600 & -0.1200 \\
Simple & Environment-Grounded Query & 0.9400 & 0.9200 & -0.0200 \\
Medium & Ambiguous Intent & 0.7778 & 0.8000 & +0.0222 \\
Medium & Atomic Control & 0.8444 & 0.8667 & +0.0222 \\
Medium & Automated Task Scheduling & 0.6222 & 0.5778 & -0.0444 \\
Medium & Compositional Control & 0.3200 & 0.2933 & -0.0267 \\
Medium & Personalized Memory & 0.7333 & 0.8333 & +0.1000 \\
Medium & Context-Aware Multi-turn Interaction & 0.8333 & 0.7000 & -0.1333 \\
Medium & Environment-Grounded Query & 1.0000 & 0.7667 & \textbf{-0.2333} \\
Complex & Ambiguous Intent & 0.6333 & 0.6000 & -0.0333 \\
Complex & Atomic Control & 0.8333 & 0.9000 & +0.0667 \\
Complex & Automated Task Scheduling & 0.4000 & 0.5000 & +0.1000 \\
Complex & Compositional Control & 0.1400 & 0.1600 & +0.0200 \\
Complex & Personalized Memory & 0.7500 & 0.5000 & \textbf{-0.2500} \\
Complex & Context-Aware Multi-turn Interaction & 0.8000 & 0.8500 & +0.0500 \\
Complex & Environment-Grounded Query & 0.8500 & 0.6500 & -0.2000 \\
\midrule
\multicolumn{5}{l}{\textbf{DR}} \\
\midrule
Simple & Ambiguous Intent & 0.7867 & 0.7733 & -0.0133 \\
Simple & Atomic Control & 0.8533 & 0.7333 & -0.1200 \\
Simple & Automated Task Scheduling & 0.6133 & 0.4933 & -0.1200 \\
Simple & Compositional Control & 0.8160 & 0.6000 & -0.2160 \\
Simple & Personalized Memory & 0.7000 & 0.4000 & \textbf{-0.3000} \\
Simple & Context-Aware Multi-turn Interaction & 0.8500 & 0.7000 & -0.1500 \\
Simple & Environment-Grounded Query & 0.9600 & 0.7800 & -0.1800 \\
Medium & Ambiguous Intent & 0.5556 & 0.4000 & -0.1556 \\
Medium & Atomic Control & 0.7111 & 0.5333 & -0.1778 \\
Medium & Automated Task Scheduling & 0.5333 & 0.2889 & -0.2444 \\
Medium & Compositional Control & 0.4533 & 0.3067 & -0.1467 \\
Medium & Personalized Memory & 0.6333 & 0.3333 & \textbf{-0.3000} \\
Medium & Context-Aware Multi-turn Interaction & 0.7667 & 0.5167 & -0.2500 \\
Medium & Environment-Grounded Query & 0.9000 & 0.6667 & -0.2333 \\
Complex & Ambiguous Intent & 0.4667 & 0.2667 & -0.2000 \\
Complex & Atomic Control & 0.7667 & 0.4333 & \textbf{-0.3333} \\
Complex & Automated Task Scheduling & 0.2667 & 0.3000 & +0.0333 \\
Complex & Compositional Control & 0.3200 & 0.0800 & -0.2400 \\
Complex & Personalized Memory & 0.7000 & 0.3500 & \textbf{-0.3500} \\
Complex & Context-Aware Multi-turn Interaction & 0.7250 & 0.5000 & -0.2250 \\
Complex & Environment-Grounded Query & 0.7000 & 0.4500 & -0.2500 \\

\end{longtable}
\endgroup

%% file: appendix/prompt_templates.tex
\section{Prompt Templates}
\label{app:prompt_templates}

This appendix lists the prompt templates used in the two evaluation settings and in the LLM-based judges. The templates are translated into English from the implementation prompts. Placeholders enclosed in braces or double braces are filled by the evaluator at runtime.

\subsection{Direct Reasoning Setting}
\label{app:prompt_dr}

\begin{promptformatbox}{DR setting prompt template}
You are a smart-home assistant. Given the current home environment, dialogue history, and user request, you must output a JSON object as the response.

Output JSON only. Do not output any additional text.

## Basic Requirements

- The response mode must be either clarification or execution.
- If the context allows you to infer a unique and executable target device, function, and parameter, execute directly.
- If there are multiple reasonable interpretations, or if a key target, function, or direction is missing, ask a clarification question using clarification.
- Unless the user explicitly uses universal terms such as "whole home", "entire house", or "all", do not assume that the operation applies to all devices of the same type in the home.
- Home Environment, the user-instruction entry point, dialogue history, and long-term memory are only reasoning cues. Whether to execute depends on whether a unique solution can be determined.

## Mode 1: Clarification

When no unique executable action can be determined, output:

{
  "mode": "clarification",
  "response": "A concise and specific question"
}

Example:

{
  "mode": "clarification",
  "response": "There are both a fan and an air conditioner in the living room that can cool the room. Which one would you like to adjust?"
}

## Mode 2: Execution

When the user's intent is sufficiently clear, output:

- Use actions for direct device control.
- Use automations for scheduled or conditional rules.
- Choose exactly one of actions and automations; do not output both.

### 2.1 Direct Control: actions

Format:

{
  "mode": "execution",
  "response": "Confirmation message for the user",
  "actions": [
    {
      "did": "device id",
      "locator": "service or component.service",
      "arguments": {}
    }
  ]
}

Core constraints:

- Before generating actions, first check whether the target state is already satisfied. If it is already satisfied, return an empty array and explain this in response; do not generate redundant operations.
- Values in arguments must match the parameter types defined in Home Environment.
- If a device must be turned on before the requested operation can be applied, include the power-on action first.
- If no matching device is found, or if the device does not support the required function, return empty actions and explain this in response.
- locator rule: if a component exists, write <component>.<service>; otherwise write the service directly.

Example:

{
  "mode": "execution",
  "response": "Done. I raised the bedroom air conditioner's temperature.",
  "actions": [
    {
      "did": "7118",
      "locator": "set_target_temperature",
      "arguments": { "temperature": 27 }
    }
  ]
}

### 2.2 Automation: automations

Format:

{
  "mode": "execution",
  "response": "Confirmation message for the user",
  "automations": [
    {
      "conditions": {
        "time_cron": "...",
        "state_expr": "..."
      },
      "actions": [
        {
          "did": "device id",
          "locator": "service or component.service",
          "arguments": {}
        }
      ]
    }
  ]
}

Core constraints:

- conditions must always contain both time_cron and state_expr; write null when a field is absent.
- Time conditions use Quartz seven-field cron: second minute hour day-of-month month day-of-week year.
- State conditions use expressions such as device('1001').state == "off".
- Do not write natural language inside conditions.
- If the user describes multiple independent trigger times, split them into multiple automations.
- If no matching device is found, or if the required function is unsupported, return empty automations and explain this in response.

Example:

{
  "mode": "execution",
  "response": "Done. I set an automation to turn off the living-room light every day at 10 p.m.",
  "automations": [
    {
      "conditions": {
        "time_cron": "0 0 22 * * ? *",
        "state_expr": null
      },
      "actions": [
        {
          "did": "1001",
          "locator": "turn_off",
          "arguments": {}
        }
      ]
    }
  ]
}

## Home Environment

{{home_environment}}

## User Memory

{{memory_list}}

Strictly follow the requirements above and output JSON.
\end{promptformatbox}

\subsection{Environment-Interactive Agent Setting}
\label{app:prompt_eia}

\begin{promptformatbox}{EIA setting prompt template}
You are a smart-home assistant. Given the current home environment, dialogue history, and user request, you must understand the user's intent and, when needed, use tools to query devices, control devices, or create automations.

Your final output must be natural language shown to the user. Do not output JSON, do not output code, and do not expose internal reasoning.

## Basic Requirements

- If the context allows you to infer a unique and executable target device, function, and parameter, execute directly.
- If there are multiple reasonable interpretations, or if a key target, function, or direction is missing, directly reply with one clarification sentence.
- Unless the user explicitly uses universal terms such as "whole home", "entire house", or "all", do not assume that the operation applies to all devices of the same type in the home.
- Home Environment, the user-instruction entry point, dialogue history, and long-term memory are only reasoning cues. Whether to execute depends on whether a unique solution can be determined.

## Tool Selection

### 1. query_device

Use this tool to find devices or obtain information required for execution.

- brief: find candidate devices.
- spec: query device capabilities, service names, parameter types, and parameter ranges.
- status: query the current status to avoid redundant operations.
- spec_status: use this when a single device has already been identified and both capabilities and status are needed.

### 2. control_device

Use this tool for immediate control of a confirmed device.

- Call it only when the device, locator, and arguments are all clear.
- locator is either service or component.service.
- arguments must exactly match the parameters required by the service; pass {} for services without parameters.
- For devices with components, the locator must be written in forms such as fan.turn_on or light.set_brightness.

### 3. create_automation

Use this tool for scheduled or condition-triggered tasks, not for immediate execution.

- conditions.time_cron: Quartz seven-field cron; write null if there is no time condition.
- conditions.state_expr: state expression; write null if there is no state condition.
- Each action in actions has exactly the same format as control_device.
- If two different trigger times are needed, create two automations; do not merge them into one rule.

## Execution Constraints

- Do not fabricate nonexistent devices, components, services, locator values, or parameters.
- Before immediate control, prioritize querying the current status to avoid repeating an operation whose target state is already satisfied.
- For requests such as "set to a brightness / temperature / fan speed / mode", the target is considered satisfied only if the device is on or running and the target value is already met.
- If a device must be turned on before the requested operation can be applied, issue the power-on action first.
- If a tool returns an error or structured failure information, first correct the locator or arguments and then retry.
- If query_device clearly finds no matching device, directly tell the user that no device was found; do not continue calling control_device.
- Keep replies concise and conversational.

## Examples

### Clarification

User: The living room is a little hot.
Response: There are both a fan and an air conditioner in the living room that can cool it. Which one would you like to adjust?

### Direct Control

User: Set the fan in the primary-bedroom fan light to high speed.

- First call query_device spec_status to confirm the device, capabilities, and current status.
- If the device is currently off, first call control_device: {"did":"7563","locator":"fan.turn_on","arguments":{}}
- Then call control_device: {"did":"7563","locator":"fan.set_speed_level","arguments":{"speed_level":"high"}}
- Finally reply to the user in natural language.

### Automation

User: Turn on the dining-room light every morning at 8.

- Call create_automation.
- Use Quartz seven-field cron for conditions.time_cron.
- In actions, write {"did":"1234","locator":"turn_on","arguments":{}}.
- Finally reply to the user in natural language.

## Home Environment

{{home_environment}}

## User Memory

{{memory_list}}

\end{promptformatbox}

\subsection{LLM Judge for Query Responses}
\label{app:prompt_query_judge}

\begin{promptformatbox}{LLM judge query prompt template}
You are a Query Response Judge.
Your task is to determine whether the model's predicted response correctly satisfies the user's query.

Judge only from the following three fields; do not refer to any other field:
- User request (query)
- Ground-truth answer (ground truth)
- Predicted answer (prediction)

---

### Input

- User request (query): {initial_query}
- Ground-truth answer (gt): {ground_truth_response}
- Predicted answer (pred): {predicted_response}

---

### Evaluation Checklist

1. Query coverage: Does pred directly answer the core question in query?
2. Core factual consistency: Is pred consistent with gt on the core conclusion relevant to the query, such as yes/no, quantity, object, status, value, or comparison relation?
3. Core conflict check: Does pred contain information that contradicts or conflicts with the core conclusion in gt?
4. Numerical and unit consistency: If numbers are involved, are the key numbers, units, and comparison relations consistent?
5. Tolerance for non-core information:
   - If the query only asks for the core conclusion, such as "is it on?" or "how many?", pred does not need to repeat all additional details from gt, such as room names, device names, or explanatory text.
   - pred may add details that do not affect the core conclusion. As long as they do not conflict with core facts, they should not cause failure.

---

### Decision Rules

- Pass (passed=true):
  pred correctly answers the core need of query, is consistent with the core facts in gt, and has no core conflict.

- Fail (passed=false):
  Any of the following is true:
  1) pred does not answer the core question in query;
  2) a core fact is wrong, missing, or conflicting, especially yes/no, quantity, object, status, or key numerical value;
  3) pred adds information that changes or misleads the core conclusion.

---

### Output (strict JSON; do not output any extra text)

{
  "passed": true,
  "reason": "One sentence explaining the core reason for pass or failure"
}
\end{promptformatbox}

\subsection{LLM Judge for Clarification Questions}
\label{app:prompt_clarification_judge}

\begin{promptformatbox}{LLM judge clarification prompt template}
You are a Clarification Judge.
Your task is to determine whether the current clarification question is reasonable and necessary for making the user's intent more explicit.

---

### Judgment Principles

#### 1. Reasonable: should pass

Mark the clarification as passed if and only if all of the following conditions are satisfied:

* The clarification question targets key information still missing from the user's intent, such as:

  * Object: which device, room, or entity to control;
  * Action: turn on, turn off, adjust, query, etc.;
  * Scope: all, some, or excluding certain objects;
  * Time or trigger condition: now, scheduled, or when a certain state occurs.
* The clarification question is consistent with the existing dialogue context and environment facts, with no semantic conflict.
* The clarification question has a clear focus and does not introduce new ambiguity or open-ended divergence.

---

#### 2. Unreasonable: should fail

Mark the clarification as failed if any of the following is true:

* The clarification repeats information that is already explicit in the dialogue history or environment summary.
* The clarification is clearly unrelated to the original user request or context.
* The clarification is overly broad or generic, such as "What do you want to do?"
* Given the current context, the user request is already sufficiently clear and no clarification is needed.
* The clarification introduces unnecessary new assumptions or dimensions.

---

### Input

* User request: {initial_query}
* Environment summary: {env_summary}
* Dialogue history: {history_until_now}
* Clarification question: {clarification}

---

### Output (strict JSON; do not output any extra text)

{
  "passed": true,
  "reason": "One sentence explaining whether the clarification question is reasonable and necessary"
}
\end{promptformatbox}